\documentclass[lettersize,journal]{IEEEtran}
\usepackage{amsmath,amsfonts}
\usepackage{algorithmic}
\usepackage{algorithm}
\usepackage{array}
\usepackage[caption=false,font=normalsize,labelfont=sf,textfont=sf]{subfig}
\usepackage{textcomp}
\usepackage{stfloats}
\usepackage{url}
\usepackage{verbatim}
\usepackage{graphicx}
\usepackage{cite}
\usepackage{multirow}
\usepackage{multicol}
\usepackage{enumitem}
\usepackage{algorithm}
\usepackage{algorithmic}
\usepackage{bm}
\usepackage{pifont}
\usepackage{hyperref}
\usepackage{booktabs}
\usepackage{mathrsfs}

\usepackage{hyperref}       
\usepackage{url}            
\usepackage{booktabs}       
\usepackage{amsfonts}       
\usepackage{nicefrac}       
\usepackage{microtype}      
\usepackage{xcolor}         
\usepackage{algorithmic}
\usepackage{algorithm}
 \usepackage{amsmath} 
\usepackage{multirow}
\usepackage{booktabs}
\usepackage{longtable}
\usepackage{bbding} 
\usepackage{graphicx}

\usepackage{wrapfig}

\usepackage[numbers]{natbib}
\usepackage{xcolor}
\usepackage[normalem]{ulem}
\usepackage{soul} 
\usepackage{booktabs}

\usepackage{amsmath, amssymb, amsthm}

\newtheorem{hypothesis}{Hypothesis}

\begin{document}

\title{Text Adversarial Attacks with Dynamic Outputs}

\author{Wenqiang Wang,
        Siyuan Liang,
        Xiao Yan,
     
        Xiaochun Cao,~\IEEEmembership{Senior~Member, ~IEEE}
}


\markboth{Journal of \LaTeX\ Class Files,~Vol.~14, No.~8, August~2024}%
{Shell \MakeLowercase{\textit{et al.}}: A Sample Article Using IEEEtran.cls for IEEE Journals}

\IEEEpubid{0000--0000/00\$00.00~\copyright~2020 IEEE}


\maketitle
\begin{abstract}

Text adversarial attack methods are typically designed for static scenarios with fixed numbers of output labels and a predefined label space, relying on extensive querying of the victim model (query-based attacks) or the surrogate model (transfer-based attacks). 
However, real-world applications often involve non-static outputs.
Large Language Models (LLMs), for instance, may generate labels beyond a predefined label space, and in multi-label classification tasks, the number of predicted labels can vary dynamically with the input. We refer to this setting as the \textit{Dynamic Outputs (DO)} scenario. Existing adversarial attack methods are not designed for such settings and thus fail to effectively generate adversarial examples under DO conditions.

To address this gap, we introduce the \textit{Textual Dynamic Outputs Attack (TDOA)} method, which employs a clustering-based surrogate model training approach to convert the DO scenario into a static single-output scenario. 
To improve attack effectiveness, we propose the \textit{farthest-label targeted attack} strategy, which selects adversarial vectors that deviate most from the model’s coarse-grained labels, thereby maximizing disruption.
We extensively evaluate TDOA on four datasets and eight victim models (e.g., ChatGPT-4o, ChatGPT-4.1), showing its effectiveness in crafting adversarial examples and its strong potential to compromise LLMs with limited access.
With a single query per text, TDOA achieves a maximum attack success rate of 50.81\%. Additionally, we find that TDOA also achieves SOTA performance in conventional static output scenarios, reaching a maximum ASR of 82.68\%.
Meanwhile, by conceptualizing translation tasks as classification problems with unbounded output spaces, we extend the TDOA framework to generative settings, surpassing prior results by up to 0.64 RDBLEU and 0.62 RDchrF.

\end{abstract}


\begin{figure*}\label{fig_overview1}
  \centering
  \includegraphics[width=1\textwidth]{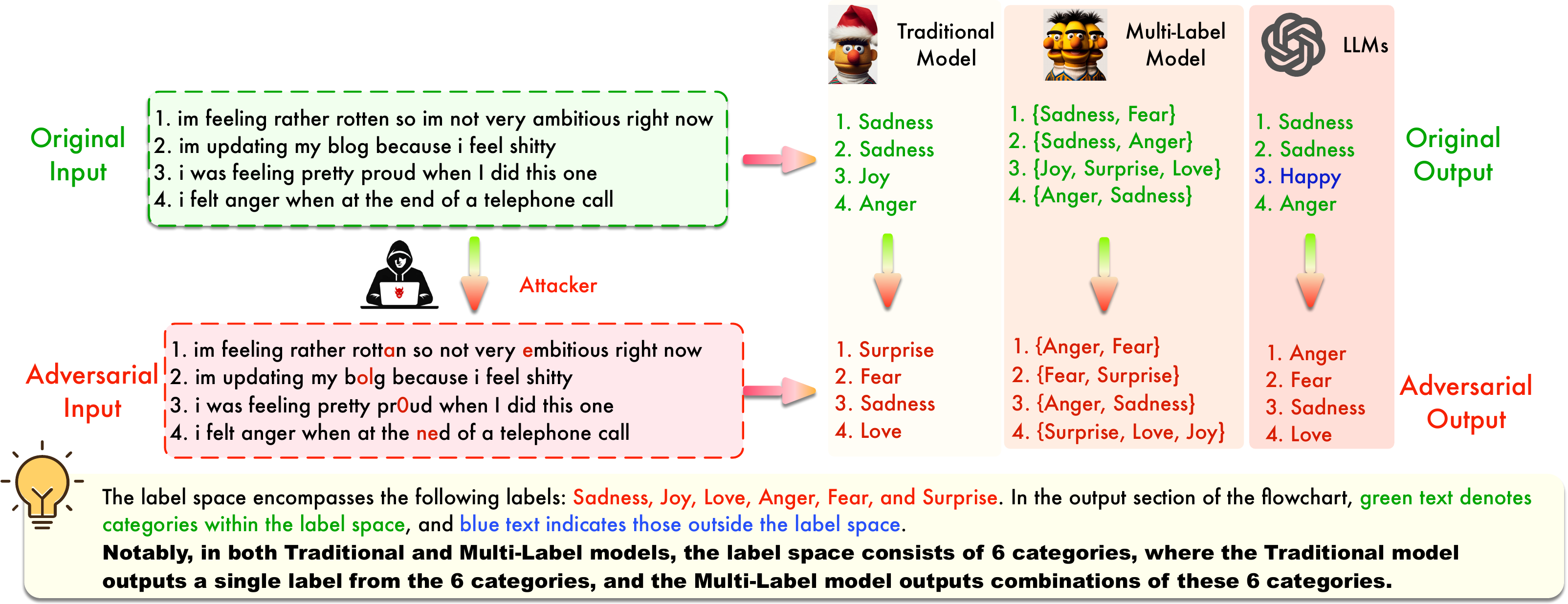} 
  \caption{The overview of \textbf{Dynamic Outputs (DO)} scenario. ``Traditional'' indicates the traditional static output scenario. 
  ``multi-label'' indicates the multi-label classification task in DO scenario, the label number changes with different input texts. ``LLMs'' indicates the text classification by
  LLMs in DO scenario, the output label may exceed the label space.}\label{Overview-DO}
  \vspace{-0.1in}
\end{figure*}
\section{Introduction}
\label{sec:intro}
Text classification is a fundamental component in text-related tasks. However, text classification models are susceptible to adversarial attacks, where minor perturbations to the words can lead to substantial changes in the model's output.
An adversarial example~\cite{wei2018transferable,liang2020efficient,liang2022large,liang2022parallel,muxue2023adversarial,wang2023diversifying,liu2023x} is a carefully crafted input that is often nearly indistinguishable from the original input and is intentionally modified to cause a model to produce an unexpected output.
In current research on adversarial text classification attacks, the victim model consistently generates the \textit{fixed number of labels} and the \textit{static output space}~\cite{waghela2024modified,han2024bfs2adv,zhu2024limeattack,kang2024hybrid}. In this setting, predictions are restricted to the predefined ground-truth label set, meaning that the model always outputs a static label.

In contrast, real-world applications often deviate from this assumption. As illustrated in Fig.~\ref{Overview-DO}, multi-label classification tasks allow the number of predicted labels to vary depending on the input text~\cite{kassim2024multi, fan2024learning}. 
For example, in a static-output setting, Sentence~2 may be labeled only as ``\texttt{Sadness}''. However, in a multi-label classification setting, the same sentence could be labeled as both ``\texttt{Sadness}'' and ``\texttt{Anger}''. Similarly, Sentence~3 may be labeled as ``\texttt{Joy}'' in the static-output case, but as ``\texttt{Joy}'', ``\texttt{Surprise}'', and ``\texttt{Love}'' in the multi-label case. These examples highlight that the number of output labels is \textbf{dynamic}, varying with the input text.

Beyond multi-label tasks, Large Language Models (LLMs) introduce an additional source of output variability. Unlike conventional discriminative classifiers, LLMs are generative models and can output labels outside the predefined ground-truth label space~\cite{fan2024learning}. For instance, when instructed to classify texts using the label set {``\texttt{Sadness}'', ``\texttt{Joy}'', ``\texttt{Love}'', ``\texttt{Anger}'', ``\texttt{Fear}'', and ``\texttt{Surprise}''}, an LLM may output ``\texttt{Happy}'' for Sentence~3, a label not included in the predefined set.
We define this phenomenon as the \textbf{dynamic output (DO)} scenario, where either the number of predicted labels or their content varies depending on the input text.

Current text classification attack methods generally assess word importance either by querying the victim model (query-based attack) or using the surrogate model (transfer-based attack). These methods sequentially perturb important words until the predicted label with the highest probability from the victim model changes.  However, these methods prove ineffective in a DO scenario, as they rely on a single predicted label and are unable to accommodate multiple or dynamic output labels. Before presenting specific attack methods, we first ensure that our approach is adaptable to the more realistic conditions of the DO scenario, which includes the \textit{limited queries constraint}, and \textit{hard-label black-box constraint}. 
Specifically, we  design more realistic attack scenarios, with their key constraints outlined as follows:
 \ding{182} DO scenario constraint: In this scenario, the number of labels varies with different input texts, and the content of these labels may extend beyond the ground-truth labels.
\ding{183} Limited queries constraint: Attackers are limited to a finite number of queries for each input text, reflecting real-world constraints on resources and the risk of detection.
\ding{184} Hard-label black-box constraint: In this scenario, attackers only have access to \textit{hard-label} outputs (i.e., the final classification label), without any gradient or probabilistic data, emulating real-world hard-label black-box attacks.

\enlargethispage{-1\baselineskip}
 
To address these challenges, 
a natural strategy is to \textbf{map the DO scenario to a static-output scenario}, where most existing attack methods are applicable. As illustrated in Fig.~1, for Sentence~3, the multi-label model produces labels such as ``\texttt{Joy}'', ``\texttt{Surprise}'', and ``\texttt{Love}'', while the LLM outputs ``\texttt{Happy}''. Although the labels differ, they all represent ``\texttt{Positive}'' emotions. Compared to fine-grained labels like ``\texttt{Joy}'', ``\texttt{Surprise}'', and ``\texttt{Happy}'', the label ``\texttt{Positive}'' corresponds to a coarser-grained semantic category, which is inherently static. By mapping fine-grained labels to such coarse-grained categories, the dynamic output scenario can be reduced to a static one.
Since fine-grained and coarse-grained labels are semantically related, and the cluster assumption~\ref{PSA} suggests that semantically similar texts are closer in feature space, clustering methods can be effectively employed to infer coarse-grained categories from fine-grained outputs.

Based on the observation that coarse-grained labels can serve as stable semantic categories, we propose the Textual Dynamic Outputs Attack (TDOA) method, a hard-label black-box attack designed to handle variable label spaces. First, to overcome the challenge of a dynamic output space, where the number and content of output labels vary depending on the input, TDOA employs the \textit{clustering-based surrogate model training} approach to transform the dynamic output to a static output. This process involves extracting dynamic labels from the victim model, vectorizing these labels along with their corresponding input texts using publicly available pre-trained models, and generating \textit{static coarse-grained labels} through clustering. The surrogate model is then trained on these texts and static coarse-grained label pairs, effectively transforming the DO scenario into a static output scenario, simplifying the attack space. 
Second, to enhance the effectiveness of the attack, TDOA incorporates the Farthest Label Targeted Attack (FLTA) strategy. By calculating the semantic distances between cluster labels (coarse-grained labels) and input texts, TDOA identifies the semantically farthest label for each input text and prioritizes perturbations on the most vulnerable words to maximize disruption. Meanwhile, to complement the proposed attack framework, we introduce the Label Inconsistency Attack Success Rate (LI-ASR), a new evaluation metric tailored for the multi-label output scenario.

To evaluate the effectiveness of the TDOA method, we conduct experiments on seven victim models across four datasets. In the multi-label classification task, TDOA achieves a LI-ASR of over 29\%  one query for per text, outperforming the second-best method by 10\%. In the LLMs classification task, TDOA achieves an ASR of over 40\% with a query  per text for the several LLMs, including GPT family. 

Additionally, we find that TDOA also achieves SOTA performance in conventional static output scenarios, reaching a maximum ASR of XX.
Meanwhile, we conceptualize the translation task as a distinct form of dynamic scenario, characterized by an infinite output space in which the generated content varies according to the input. TDOA attains SOTA performance on translation tasks, exceeding previous best results by up to 0.64 RDBLEU and 0.62
RDchrF. These findings highlight the importance of developing more robust adversarial defenses and inspire future work on adaptive training strategies~\cite{lu2025adversarial,liu2023exploring} against dynamic text attacks.
The \textbf{contributions}  are outlined as follows:

\begin{itemize}
    \item 
    We propose a scenario involving a \textit{dynamic output space}. In this scenario, the number of labels produced by the model is variable, and the output label may exceed the label space.


    
    \item 
    We introduce TDOA, an attack method with a plug-and-play framework that transforms the dynamic output space into a conventional static one. TDOA achieves state-of-the-art performance in both dynamic and static scenarios and further demonstrates applicability to translation tasks.

    \item 
We propose \textit{LI-ASR} as an evaluation metric for attack methods in multi-label classification tasks within dynamic output scenarios, offering a more accurate assessment of their effectiveness.

\end{itemize}

\section{Related Work}\label{related-work}



\subsection{Query-based attack}
In past textual adversarial attacks, most methods are single-output scenarios~\cite{waghela2024modified,han2024bfs2adv,zhu2024limeattack,kang2024hybrid}. They focus on the way of transforming the original texts into adversarial examples, such as perturbing the important characters and words.~\cite{wang2022semattack,hu2024fasttextdodger,liu2024hqa,liu2023sspattack} And some methods also craft adversarial examples by attacking the sentence~\cite{lin2021using,xu2021grey,wang2020t3,le2020malcom}. These methods can be divided into three
categories based on the feedback of the victim model, including white-box attacks, soft-label black-box attacks, and hard-label black-box attacks.
In a white-box textual attack, attackers can obtain any information they need about the victim model. For example,
TextBugger~\cite{li2019textbugger} alters characters and words employing a greedy algorithm. 
Compared to white-box attacks, black-box scenarios are more practical, as attackers have access only to the model output.
In the textual soft-label black-box attack,  there are many attack methods aimed at perturbing the words based on output probabilities~\cite {lee2022query,li2020Bert}.
CT-GAT~\cite{li2020Bert} attacks words using a pre-trained Bert model. SememePSO~\cite{zang2020word} optimizes the search space to create adversarial examples. 
In the hard-label black-box attack,  adversarial examples are generated just based on the output labels and multiple queries~\cite{hu2024fasttextdodger,liu2024hqa,liu2023sspattack}.

\subsection{Transfer-based Attack}
A transfer-based attack is a type of black-box attack in which adversarial examples generated on a surrogate model can transfer to and successfully mislead the victim model.
~\citep{papernot2017practical,dong2018boosting}.
Then, in the absence of a surrogate model, several studies demonstrate that auxiliary data can also facilitate successful attacks through training a surrogate model and leveraging transfer attacks~\citep{li2020practical,sun2022towards}. Additionally, more effective loss functions have been proposed to train surrogate models~\citep{wang2021feature,richards2021adversarial}, as well as techniques to refine surrogate models~\citep{xiaosen2023rethinking,yuan2021meta}.

\begin{figure*}\label{fig_overview1}
  \centering
  \includegraphics[width=1\textwidth]{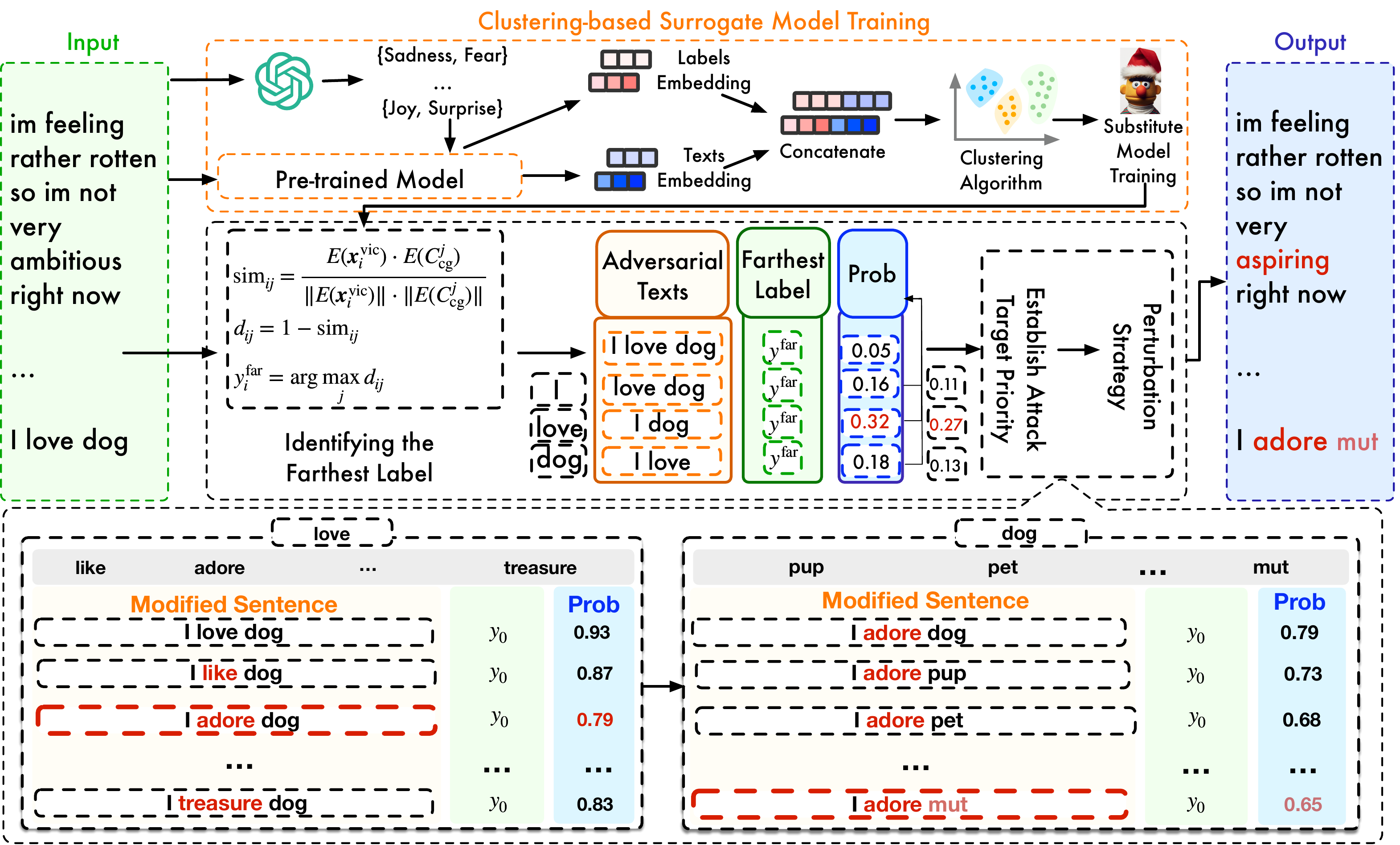} 
  \caption{The Overview of TDOA. TDOA comprises two components: (1) the training of the surrogate model utilizing coarse-grained labels derived from a clustering algorithm, and (2) the implementation of a farthest label targeted attack to enhance the attack's impact.}
  \vspace{-0.1in}
\end{figure*}

\section{Preliminary}\label{PSA}
\textbf{Multi-label classification:}
The multi-label classification poses a significant challenge in machine learning, as each instance can be associated with multiple labels, unlike traditional single-label classification, where each instance has only one. For example, articles may be categorized under multiple genres, such as ``Technology'' and ``Business''. Similarly, musical tracks may encompass multiple styles, blending genres like ``Rock'', ``Grunge'', and ``Pop''.

\textbf{Targeted attack and untargeted attack:}
\ding{182} \textit{Targeted Attack}: Force misclassification into a specific label (e.g., kitten → dog).
\ding{183} \textit{Untargeted Attack}: Cause any incorrect classification (e.g., kitten → any non-kitten).

\textbf{Transferability and cross-task transferability:}
\textit{Transferability} allows adversarial examples crafted on one (surrogate) model to remain effective on another (victim) model.  
\textit{Cross-task transferability} extends transferability across different tasks, e.g., from classification to summarization.  
\textit{Thus, adversarial examples generated using static surrogate models may effectively attack dynamic victim models.}

\textbf{Victim texts and auxiliary data:}
\textit{Victim texts} are the targets for adversarial generation.  
\textit{Auxiliary data} are external datasets (e.g., \(x_{\text{train}}, y_{\text{train}}\)) used to train surrogate models. Real-world conditions may involve few-shot, or low-quality auxiliary data, which poses additional challenges.


\textbf{Cluster and manifold assumption:}
\ding{182} \textit{Cluster Assumption:} In feature space, similarity can be quantified using distance metrics (e.g., Euclidean distance, cosine similarity). Semantically similar texts are located closer together in feature space.
\ding{183} \textit{Manifold Assumption:} High-dimensional data lie on a low-dimensional manifold, with similar instances being closer to each other on the manifold.

\section{Definition, Observation  and Motivation}
\subsection{Definition of The DO}
The input text is denoted as \( \boldsymbol{x} \), the multi-label classification model as \( f_{\text{multi}} \), and the LLMs classification process as \( f_{\text{LLMs}} \). The label space is \( \{C_1, C_2, \cdots, C_m\} \). The multi-label dynamic setting is formulated in Equation~\ref{multi_equ}.
\begin{equation}\label{multi_equ}
\small
y^1,y^2,\cdots,y^w=f_{\text{multi}}(\boldsymbol{x}), 1\leq w \leq m, ~ y \in \{C_1,C_2,\cdots,C_m\},
\end{equation}
where  \( w \) denotes the number of labels predicted by the multi-label model \( f_{\text{multi}} \), ranging from 1 to \( m \). Therefore, the number of output labels is a \textit{dynamic variable}. Meanwhile, the corresponding formulation in the LLMs setting is given in Equation~\ref{llms_equ}.
\begin{equation}\label{llms_equ}
y=f_{\text{LLMs}}(\boldsymbol{x}), y  \text{ may }  \notin \{C_1,C_2,\cdots,C_m\}.
\end{equation}
Here, \( y \) represents the label predicted by LLMs for a given input \( \boldsymbol{x} \). As generative models, LLMs may produce labels that fall outside the predefined label space \( \{C_1, C_2, \cdots, C_m\} \). Equations~\ref{multi_equ} and~\ref{llms_equ} depict two core dynamic scenarios: the variation in the number of output labels and the variation in the content of output labels, respectively. While LLM-based multi-label classification involves both, this study mainly analyzes these two basic dynamics separately.

\subsection{Observation  and Motivation}\label{Motivation}
As illustrated in Figure \ref{Overview-DO}, in the multi-label scenario, the predicted labels for Sentence 3 (``\texttt{Joy}'', ``\texttt{Surprise}'', ``\texttt{Love}'') consistently represent positive emotions. Likewise, in the LLMs scenario, the predicted label (``\texttt{Happy}'') does not appear in the predefined label space (``\texttt{Sadness}'', ``\texttt{Joy}'', ``\texttt{Love}'', ``\texttt{Anger}'', ``\texttt{Fear}'', and ``\texttt{Surprise}'') but still denotes a positive emotion. We categorize labels such as ``\texttt{Sadness}'', ``\texttt{Joy}'', ``\texttt{Love}'', ``\texttt{Anger}'', etc., as fine-grained labels, while ``\texttt{Positive}''  are coarse-grained labels.
Our findings indicate that in dynamic scenarios, while fine-grained labels are dynamic, their coarse-grained counterparts remain static. Building on this, we propose  Coarse-Grained Label Transfer Attack Hypothesis, wherein adversarial examples that modify a coarse-grained label also induce changes in its fine-grained labels. To leverage this principle, we extract coarse-grained labels from text–fine-grained label pairs and train a coarse-grained label surrogate model. Adversarial examples generated by this surrogate model effectively transfer to the fine-grained label victim model, successfully attacking the fine-grained classification system. The Coarse-grained Label Transfer Attack Hypothesis is formulated as Hypothesis \ref{Coarse-Grained_Label_Transfer_Attack}.

\begin{hypothesis}[\textbf{Coarse-grained Label Transfer Attack}]
\label{Coarse-Grained_Label_Transfer_Attack}
Adversarial examples generated by the surrogate model \( f_{\text{s}} \), which is trained with coarse-grained labels, can successfully attack both the multi-label classification victim model \( f_{\text{multi}} \) and the LLMs victim model \( f_{\text{LLMs}} \), which are trained with fine-grained labels.
 Formally:
 \begin{equation}
 \begin{array}{c}
 f_{\text{s}}(\boldsymbol{x}) \neq f_{\text{s}}(\tilde{\boldsymbol{x}}),
\text{  }  f_{\text{LLMs}}(\boldsymbol{x}) \neq f_{\text{LLMs}}(\tilde{\boldsymbol{x}}), \\ 
\text{ and }  \{f_{\text{multi}}(\boldsymbol{x}) \cap f_{\text{multi}}(\tilde{\boldsymbol{x}})\} \to \varnothing,
\end{array}
\end{equation}
where the 
\( \boldsymbol{x} \) represents the input text, while \( \tilde{\boldsymbol{x}} \) denotes its adversarial example, generated by the surrogate model \( f_{\text{s}} \). \( f_{\text{LLMs}}(\boldsymbol{x}) \neq f_{\text{LLMs}}(\tilde{\boldsymbol{x}}) \) signifies that the predict labels for \( \boldsymbol{x} \) and \( \tilde{\boldsymbol{x}} \) differ, indicating a successful attack on \( f_{\text{LLMs}} \).  
Additionally, the expression  
$
\{f_{\text{multi}}(\boldsymbol{x}) \cap f_{\text{multi}}(\tilde{\boldsymbol{x}})\} \to \varnothing
$
implies that the predicted label sets for \( \boldsymbol{x} \) and \( \tilde{\boldsymbol{x}} \) in \( f_{\text{multi}} \) gradually converge to an empty set. Thus, \( \tilde{\boldsymbol{x}} \) also successfully attacks \( f_{\text{multi}} \).
\end{hypothesis}
For instance, when Sentence 3 in Figure \ref{Overview-DO} is classified by the multi-label model, its predicted labels are ``\texttt{Joy}'', ``\texttt{Surprise}'', and ``\texttt{Love}''. In contrast, the LLMs predict only ``\texttt{Happy}''.  
If the same sentence is classified by a coarse-grained surrogate model trained on a three-class sentiment framework (``\texttt{Positive}'', ``\texttt{Neutral}'', ``\texttt{Negative}''), and its predicted label is ``\texttt{Positive}'', then generating an adversarial example that shifts the label to ``\texttt{Neutral}'' or ``\texttt{Negative}'' will also modify the fine-grained labels in both the multi-label model and LLMs.  
Consequently, the adversarial example diverges from ``\texttt{Positive}'' label, facilitating a successful attack on both the multi-label model and LLMs.

\section{Method}
\label{headings}
\subsection{Adversary's Goals and Constraints}
The adversary aims to maximize the discrepancy between the labels of the original text and its adversarial counterpart. For LLM classification, this requires \(y_{\text{adv}} \neq y_{\text{ori}}\). In multi-label classification, the objective is to minimize the label overlap, i.e., \(\arg\min (\boldsymbol{y}_{\text{adv}} \cap \boldsymbol{y}_{\text{ori}})\), where \(\boldsymbol{y}_{\text{adv}}\) and \(\boldsymbol{y}_{\text{ori}}\) denote the \textbf{label sets} of the adversarial and original texts. Formally:
\begin{equation}
\begin{cases}
y_{\text{adv}} \neq y_{\text{ori}}, \quad & \text{(LLMs Classification)}, \\[6pt]
arg\min (\boldsymbol{y}_{\text{adv}} \cap \boldsymbol{y}_{\text{ori}}), & \text{(Multi-label Classification)}.
\end{cases}
\end{equation}
Meanwhile, we define several constraints on the adversary's capabilities to simulate a realistic attack scenario: \ding{182} \textit{Model Feedback:} We establish a \textbf{black-box scenario with hard-label output}, where the attacker can not access crucial information, including model gradients, architecture, or soft labels (probabilities). \ding{183} \textit{Auxiliary Data:} The adversary can access 50 victim texts as auxiliary data. These victim texts are generally more accessible to attackers during the attack than the training data.
\subsection{The Overview of TDOA}
Drawing upon the observations and hypotheses in Section~\ref{Motivation}, we propose the Textual Dynamic Outputs Attack (TDOA), specifically designed for DO scenarios. 
As Fig.~\ref{fig_overview1}  shows, TDOA consists of two components: (i) a \textit{clustering-based surrogate model training} (Sections~\ref{pseudo-label} and~\ref{substitute-model}), which converts DO into a static setting and remains effective even under hard-label black-box conditions; and (ii) a \textit{farthest-label targeted attack} (Sections~\ref{farthest-label}, \ref{priority}, and \ref{perturb}), which enhances attack success and transferability by targeting the cluster label most semantically distant from the victim text.



\subsection{Clustering-based Surrogate Model Training}\label{Clustering-based_Surrogate_Model_Training}
\subsubsection{Normalized Static Coarse-grained Label Generation}\label{pseudo-label}
To address the DO scenario, we convert the DO scenario into a conventional static one. 
As shown in Fig.~\ref{Overview-DO} and Hypothesis~\ref{Coarse-Grained_Label_Transfer_Attack}, we employ a coarse-grained surrogate model to convert the DO scenario into a static setting. Therefore, the key challenge lies in how to construct coarse-grained labels based on auxiliary texts and their fine-grained labels. 
Coarse-grained labels provide a more abstract and generalized representation of the text and its fine-grained labels. Compared to fine-grained labels, they are fewer in number, capture higher-level semantic structures, and reveal latent patterns in the data. For this reason, we adopt cluster labels as coarse-grained labels.

The procedure for generating normalized coarse-grained labels is outlined below, using a multi-label classification setting as an example. 
(1) The victim model \(f_{\text{multi}}\) is queried with an auxiliary text \(\boldsymbol{x}_i\) to obtain \(l\) predicted dynamic labels, denoted \(y_i^1, y_i^2, \dots, y_i^l\):
\begin{equation}
y_i^1, y_i^2, \cdots, y_i^l = f_{\text{multi}}(\boldsymbol{x}_i).
\end{equation}
(2) A pre-trained model \(f_{\text{pre}}\) encodes both the auxiliary text and its predicted labels, yielding embeddings \(\boldsymbol{E}_i^{\text{text}}\) and \(\boldsymbol{E}_i^{\text{label}}\), which encode
semantic information from the auxiliary text and the dynamic
labels, respectively:
\begin{equation}
\boldsymbol{E}_i^{\text{text}} = f_{\text{pre}}(\boldsymbol{x}_i), \quad 
\boldsymbol{E}_i^{\text{label}} = f_{\text{pre}}(y_i^1, y_i^2, \cdots, y_i^l).
\end{equation}
(3) These embeddings are concatenated to form a combined representation \(\boldsymbol{E}_i\):
\begin{equation}
\boldsymbol{E}_i = Concat(\boldsymbol{E}_i^{\text{text}}, \boldsymbol{E}_i^{\text{label}}),
\end{equation}
where \(Concat\) denotes vector concatenation. This unified embedding \(\boldsymbol{E}_i\) integrates information from both the auxiliary text and its predicted dynamic labels.
Repeating this step for all \(n\) auxiliary texts gives \(\boldsymbol{E}_1, \dots, \boldsymbol{E}_n\).  

\noindent(4) A clustering function \(f_c\) assigns each \(\boldsymbol{E}_i\) a cluster label \(y_{\text{cg}}^i \in \{C_{\text{cg}}^1, \dots, C_{\text{cg}}^k\}\) ($k$ is the clustering number), forming the dataset $\boldsymbol{D}_{\text{cg}}$ that pairs auxiliary texts with their coarse-grained labels:
\begin{equation}
\boldsymbol{D}_{\text{cg}} = \left\{ \left( \boldsymbol{x}_1, y_{\text{cg}}^1 \right), \dots, \left( \boldsymbol{x}_n, y_{\text{cg}}^n \right) \right\}.
\end{equation}
Each auxiliary text is thus assigned a single coarse-grained label, enabling a static classification framework.

\subsubsection{Static Surrogate Model Training}\label{substitute-model}
Once the static coarse-grained labels are generated, the auxiliary texts and their corresponding coarse-grained labels are used to train the static text classification surrogate model \(f_s\), which converts the DO scenario into a static scenario. Specifically, the auxiliary texts serve as input text, while the coarse-grained labels serve as output labels for training the surrogate model. Formally:
\begin{equation}
\small
 \mathbf{\theta}^* = \arg\min_{\mathbf{\theta}} \frac{1}{|\boldsymbol{D}_{\text{cg}}|} \sum_{(\boldsymbol{x}_i, y_{\text{cg}}^i) \in \boldsymbol{D}_{\text{cg}}} (y_{\text{cg}}^i - f_{\text{s}}(\mathbf{\theta}; \boldsymbol{x}_i))^2, 
 f_s=f(\mathbf{\theta}^*),
\end{equation}
where \(y_{\text{cg}}^i\) denotes the coarse-grained label of \(\boldsymbol{x}_i\) and \(\mathbf{\theta}^*\) represents the optimal parameters of the surrogate model. The training procedure is summarized in Algorithm~\ref{A2}, following the methods of Sections~\ref{pseudo-label} and~\ref{substitute-model}.

\begin{algorithm}[ht]
	\renewcommand{\algorithmicrequire}{\textbf{Input:}}
	\renewcommand{\algorithmicensure}{\textbf{Output:}}
	\caption{Clustering-based Surrogate Model Training } \label{A2}
	\label{alg3} 
	\begin{algorithmic}
            
		\REQUIRE The auxiliary texts $\boldsymbol{D}=\left\{\boldsymbol{x}_{1}, \boldsymbol{x}_{2},  \cdots, \boldsymbol{x}_{n}\right\}$, the pre-trained model \(f_{\text{pre}}\), the clustering function  $f_{c}$, the victim  model $f_{\text{multi}}$.
		\ENSURE The surrogate model $f_s$
            \FOR{$i = 1$ to $n$}
                \STATE $y_i^1, y_i^2, \cdots, y_i^l = f_{\text{multi}}(\boldsymbol{x}_i), \boldsymbol{E}_i^{\text{text}} = f_{\text{pre}}(\boldsymbol{x}_i),$
                \STATE $\boldsymbol{E}_i^{\text{label}} = f_{\text{pre}}(y_i^1, y_i^2, \cdots, y_i^l), \boldsymbol{E}_i = Concat(\boldsymbol{E}_i^{\text{text}}, \boldsymbol{E}_i^{\text{label}}).$
            \ENDFOR
            \STATE $\boldsymbol{E}_{all}=[\boldsymbol{E}_1,\boldsymbol{E}_2,\cdots,\boldsymbol{E}_n]$
            \FOR{$i=1$ to $n$}
            \STATE Input $\boldsymbol{E}_{i}$ into the $f_c$ to generate the corresponding coarse-grained label $y_{\text{cg}}^i$,
    $y_{\text{cg}}^i=f_c(\boldsymbol{E}_{i})$.
            \ENDFOR
            \STATE The victim text coarse-grained label  pairs   $\boldsymbol{D}_{\text{cg}}=\left\{\left(\boldsymbol{x}_{1},y_{\text{cg}}^1\right),\left(\boldsymbol{x}_{2},y_{\text{cg}}^2\right), \cdots,\left(\boldsymbol{x}_{n},y_{\text{cg}}^n\right)\right\}$

            \STATE Train the surrogate model $f_s$ on $\boldsymbol{D}_{\text{cg}}$. 
            \STATE $\boldsymbol{\theta}^* = \arg\min_{\boldsymbol{\theta}} \frac{1}{|\boldsymbol{D}_{\text{cg}}|} \sum_{(\boldsymbol{x}_i, y_{\text{cg}}^i) \in \boldsymbol{D}_{\text{cg}}} \left( y_{\text{cg}}^i - f_{\text{s}}(\boldsymbol{\theta}; \boldsymbol{x}_i) \right)^2$

            \RETURN The  surrogate model $f_s=f(\boldsymbol{\theta}^*)$
            
	\end{algorithmic} 
\end{algorithm}



\subsection{Farthest Label Targeted Attack}
\begin{figure}[h]\label{cluster_label_coarese_label}
  \centering
  \includegraphics[width=1\columnwidth]{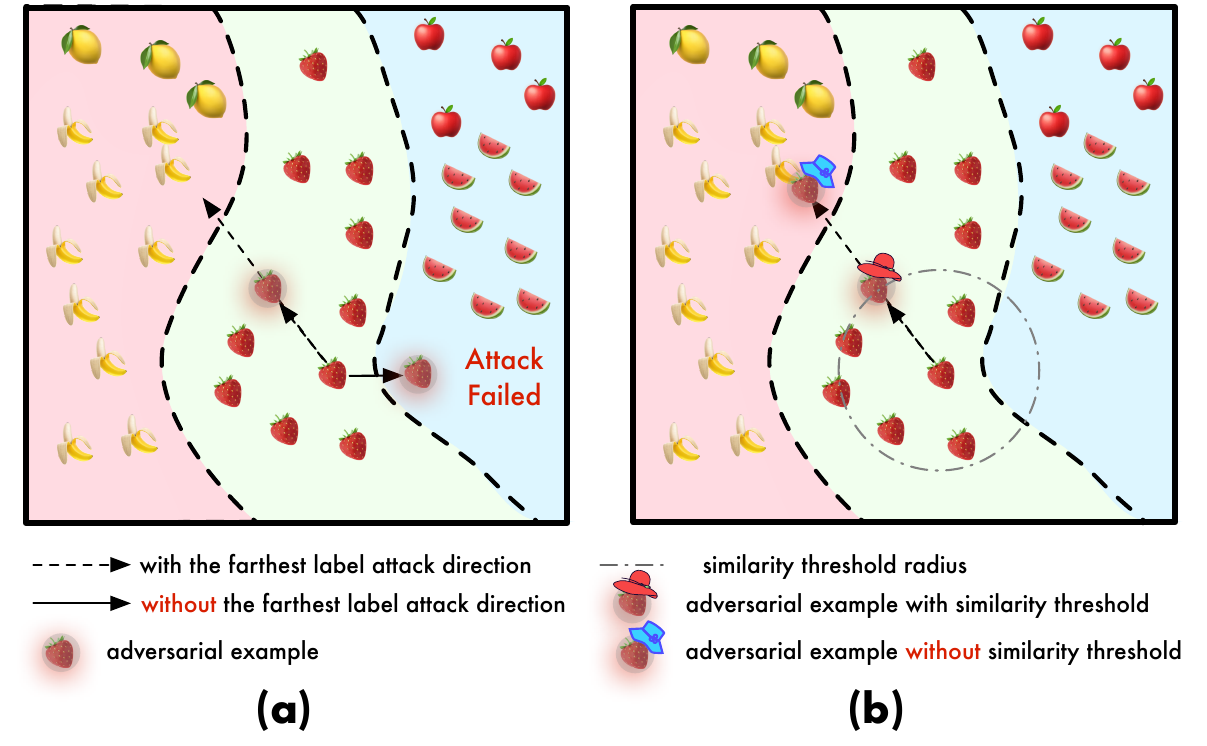} 
  \caption{Illustration of attack scenarios: (a) with and without the Farthest-Label Targeted Attack; (b) with and without the similarity threshold.
}
  \vspace{-0.1in}
\end{figure}

In Section \ref{Clustering-based_Surrogate_Model_Training}, we use cluster labels as an estimate of coarse-grained categories, which may introduce errors. For example, in a multi-label dynamic scenario, a text predicted as ``\texttt{Disappointment}'', ``\texttt{Annoyance}'', ``\texttt{Remorse}'', and ``\texttt{Surprise}'' may be clustered into the negative-emotion category. Adversarial examples generated from a surrogate model trained on this coarse label might be classified as non-negative by the surrogate model; however, since ``\texttt{Surprise}'' is also non-negative, it may still be predicted by the victim model, causing the attack to fail. Furthermore, as shown in Subfigure (a) of Fig.~\ref{cluster_label_coarese_label}, 
victim texts that lie near the decision boundary of the surrogate model may cause conventional transfer-based attacks to terminate once the predicted labels of the original and adversarial texts diverge on the surrogate. In such cases, the generated adversarial examples may still fail to mislead the victim model. Therefore, it is necessary to identify a more suitable perturbation direction for these victim texts.




To address these issues, we propose Farthest-Label Targeted Attack. Based on the cluster and manifold assumptions (Section~\ref{PSA}), semantically related fine-grained labels remain spatially close even across clusters. Leveraging this, our method targets the coarse-grained label most distant from the victim text, termed the \textit{farthest label}. The attack proceeds in three stages: \ding{182} \textit{Identifying the Farthest Label} (Section~\ref{farthest-label}); \ding{183} \textit{Establishing Target Priorities} (Section~\ref{priority}); and \ding{184} \textit{Formulating a Perturbation Strategy} (Section~\ref{perturb}).

\subsubsection{Identifying the Farthest Label}\label{farthest-label}
For each victim text $x_i^{\text{vic}}$,  we denote its farthest coarse-grained label as $y^{\text{far}}_i$. Then we perturb $x_i$ until its predicted label of $f_s$ change or the semantic similarity between the original text and the attacked text is less than $\tau$. 
We achieve this process by finding the farthest label
 through the following steps:
\ding{182} \textit{Coarse-grained label Embedding Representation.} 
The mean of the auxiliary text embeddings associated with the coarse-grained label \(C_{\text{cg}}^i \) serves as the embedding representation of \( C_{\text{cg}}^i \): 
\begin{equation}
\begin{array}{c}
\forall~ y_j = C_{\text{cg}}^i, \quad \boldsymbol{x}_j \in \textbf{S}_{C_{\text{cg}}^i}, n_i=Count(\textbf{S}_{C_{\text{cg}}^i})\\
\boldsymbol{E}({C_{\text{cg}}^i}) = \frac{1}{n_i} \sum_{\boldsymbol{x}_j \in \textbf{S}_{C_{\text{cg}}^i}} \boldsymbol{E}(\boldsymbol{x}_j), 
\end{array}
\end{equation}
where \( \textbf{S}_{ C_{\text{cg}}^i} \) denotes the set of auxiliary texts labeled with \( C_{\text{cg}}^i \),  and \( \boldsymbol{E}(\boldsymbol{x}_j) \) is the embedding of the auxiliary text \( \boldsymbol{x}_j \). The set of embeddings for all coarse-grained labels is denoted as
$\{\boldsymbol{E}({ C_{\text{cg}}^1}), \boldsymbol{E}({C_{\text{cg}}^2}), \cdots, \boldsymbol{E}({C_{\text{cg}}^k})\}$.
\ding{183} \textit{Semantic Distance Calculation.} 
We employ the cosine similarity to compute the semantic distance. 
Specifically, for each victim text \( \boldsymbol{x}_i^{\text{vic}} \), we first transform it into a vector representation and subsequently compute its cosine similarity with each coarse-grained label. Formally, this process can be expressed as:
\begin{equation}\label{sim}
\text{sim}_{ij} = \frac{\boldsymbol{E}(\boldsymbol{x}_i^{\text{vic}}) \cdot \boldsymbol{E}(C_{\text{cg}}^j)}{\|\boldsymbol{E}(\boldsymbol{x}_i^{\text{vic}})\| \cdot \|\boldsymbol{E}(C_{\text{cg}}^j)\|},
d_{ij} = 1 - \text{sim}_{ij}.
\end{equation}
\ding{184} \textit{Identifying the Farthest Coarse-grained Label.}
For each victim text $\boldsymbol{x}_i^{\text{vic}}$, we identify its farthest coarse-grained label by 
\begin{equation}
y^{\text{\text{far}}}_i = \arg\max_{j} d_{ij}.
\end{equation}


\subsubsection{Establish Attack Target Priority}\label{priority}

We propose an adversarial attack strategy based on the farthest label, which identifies optimal word targets by quantifying the probability increase of the farthest coarse-grained label. Words whose modification substantially elevates this probability are deemed ideal attack words. By perturbing such words, the surrogate model’s prediction is directed toward \( y^{\text{far}}_i \), where \( y^{\text{far}}_i \) denotes the farthest coarse-grained label associated with $\boldsymbol{x}_i^{\text{vic}}$.
Let the victim text \( \boldsymbol{x}_i^{\text{vic}} \) be a sentence \( \boldsymbol{S}_i = [w_0, \cdots, w_i, \cdots, w_u] \), where \( w_i \) is the \( i \)-th word and \( u \) is the sentence length.

\begin{figure}[h]
    \centering
        \includegraphics[width=0.9\columnwidth]{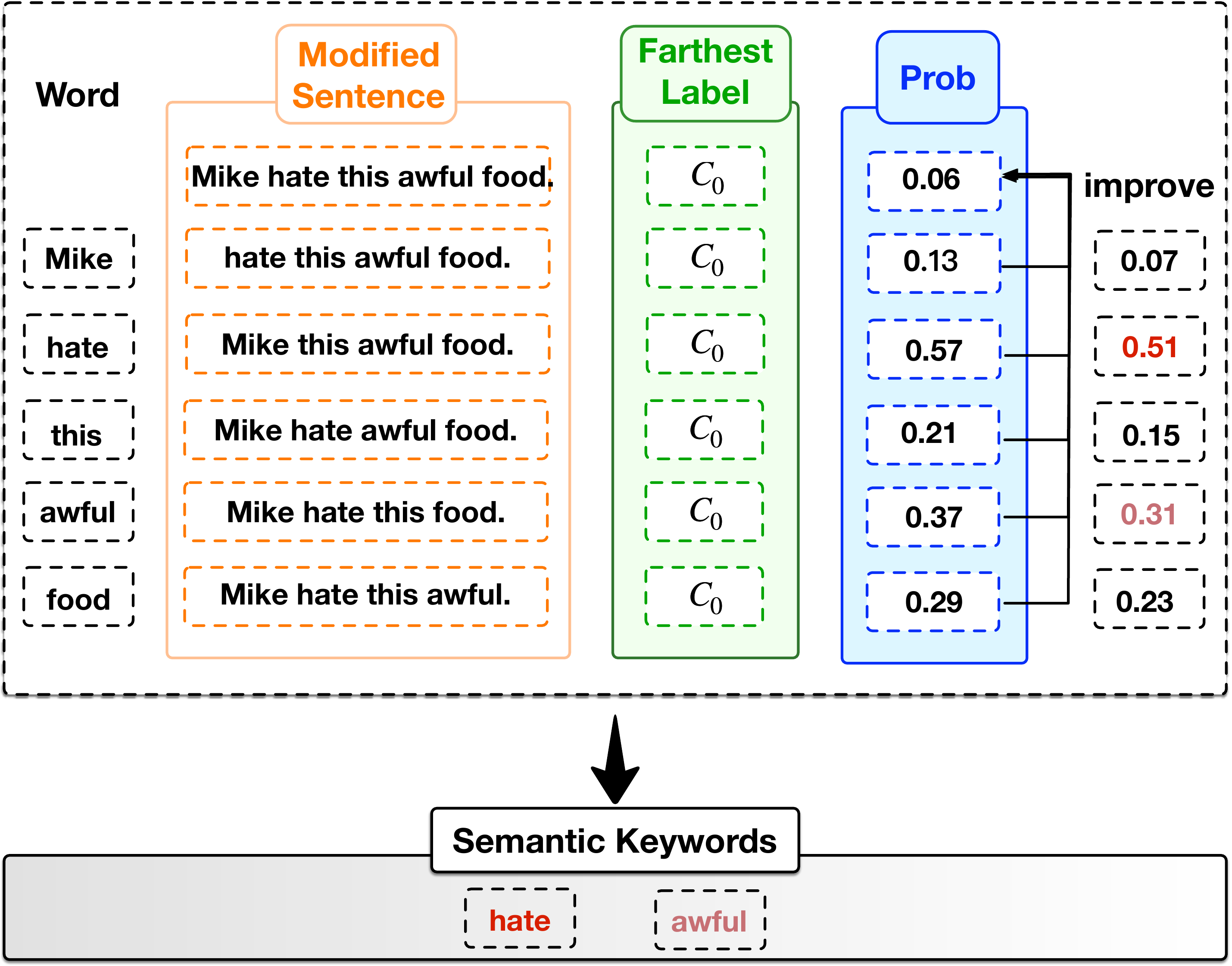}

\caption{An example of \textit{Establish Attack Target Priority}. The victim text is ``\texttt{Mike hate this awful food.}'', with the farthest label being \(C_0\), assigned a probability of 0.06 by the surrogate model. After removing ``\texttt{hate}'' and ``\texttt{awful}'', the probability of the farthest label changes most significantly; thus, these two words are prioritized for attack.
}\label{Establish_Attack_Target_Priority}
\end{figure}


Fig.~\ref{Establish_Attack_Target_Priority} illustrates the process of establishing attack target priority when the farthest label is $C_0$. 
In general, the probability assigned by the coarse-grained surrogate model $f_s$ to the farthest label $y^{\text{far}}_i$ for the victim text $\boldsymbol{x}_i^{\text{vic}}$ is relatively low (e.g., 0.06 in Fig.~\ref{Establish_Attack_Target_Priority}), denoted as $P_{f_s}(y^{\text{far}} \mid \boldsymbol{S}_i)$. 
If a word substantially influences this farthest label, removing it should increase the predicted probability. 
Accordingly, we define a variant of the sentence as 
 \( \boldsymbol{S}_{i}^{w_i} = [w_0, \cdots, w_{i-1}, w_{i+1}, \cdots, w_u] \), where the $i$-th word $w_i$ is omitted.
To prioritize attack targets, we define a priority score \( PS_{w_i} \) as:
\begin{equation}\label{importance}
\begin{split}
PS_{w_i}=P_{f_s}(y^{\text{far}} \mid \boldsymbol{S}^{w_i}_i) 
- P_{f_s}(y^{\text{far}} \mid \boldsymbol{S}_i) ,
\end{split}
\end{equation}
which measures the impact of perturbing \( w_i \) on the predicted probability of \( y^{\text{far}}_i \). 
The larger the priority score $PS_{w_i}$, the greater the likelihood that the word $w_i$ will be prioritized for attack. By ranking all words in a sentence according to their $PS$ scores (e.g., in Fig.~\ref{Establish_Attack_Target_Priority}, the words \texttt{hate} and \texttt{awful} are the first two selected words for attack), we derive the attack order for the sentence. Words that substantially increase the probability of the farthest label are regarded as stronger attack targets, thereby improving attack efficiency.

\subsubsection{Perturbation Strategy}\label{perturb}
\begin{figure}[h]
    \centering
        \includegraphics[width=\columnwidth]{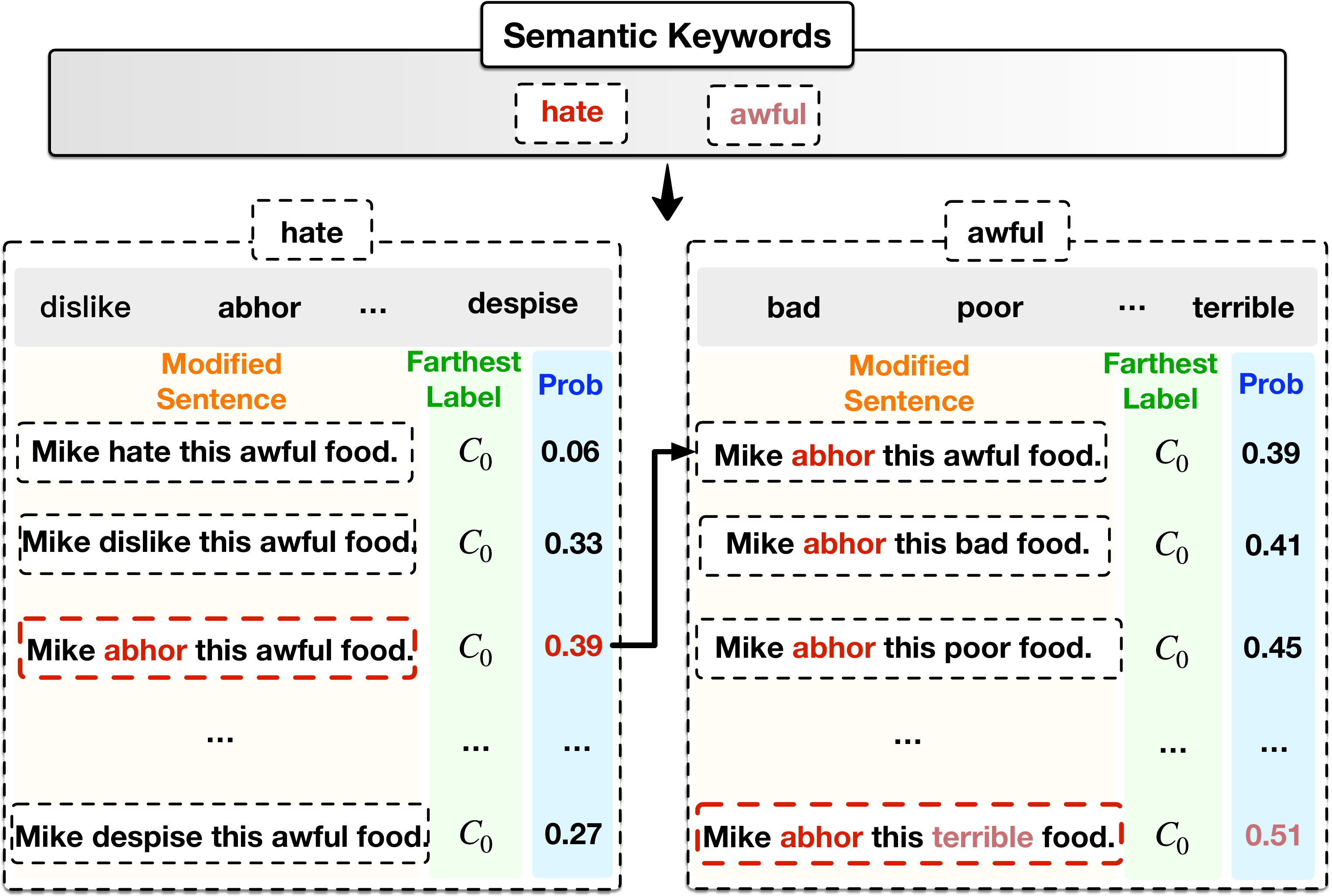}

\caption{Example of the perturbation strategy. 
The word \texttt{hate} (highest PW score) is perturbed via synonym substitution, 
with \texttt{abhor} chosen as the replacement as it yields the largest increase in the predicted probability of the farthest label. 
The word \texttt{awful} (second-highest PW score) is then perturbed using the same strategy.}
\label{Perturbation_Strategy}
\end{figure}

After determining the attack order over words in a given input \( \boldsymbol{x}_i^{\text{vic}} \), we iteratively generate perturbed candidates \( \boldsymbol{x}_i^t \) by modifying the words with the
top \( t \) \( PS \) scores. At each step, we check whether the candidate satisfies either of the following stopping criteria: 
(i) the surrogate model \( f_s \) assigns a different label to the original input; or 
(ii)  the semantic similarity between \( \boldsymbol{x}_i^t \) and \( \boldsymbol{x}_i^{\text{vic}} \) drops below a similarity threshold \( \tau \).
As shown in Subfigure (a) of Fig.~\ref{cluster_label_coarese_label}, without a similarity constraint, adversarial examples may be excessively perturbed, resulting in substantial semantic deviation. Therefore, we incorporate a similarity threshold as an additional stopping criterion for the attack.
Formally, the final adversarial example \( \tilde{\boldsymbol{x}} \) is defined as:
\begin{equation}\label{stop_equ}
\scriptsize
\tilde{\boldsymbol{x}}_i^{\text{vic}} = \boldsymbol{x}_i^{t^*}, ~ \mathrm{s.t.} ~
t^* = \min\!\left\{ t \in \mathbb{N} \,\middle|\, 
\operatorname{sim}(\boldsymbol{x}_i^t, \boldsymbol{x}_i^{\text{vic}}) \le \tau
\ \text{or} \ 
f_s(\boldsymbol{x}_i^t) \ne f_s(\boldsymbol{x}_i^{\text{vic}}) \right\}.
\end{equation}
Here,
\( \boldsymbol{x}_i^t \) is the perturbed version of \( \boldsymbol{x}_i^{\text{vic}} \) after modifying \( t \) words. \( \operatorname{sim}(\cdot, \cdot) \) is a semantic similarity function. And \( t^* \) is the minimal number of modifications required to satisfy the stopping condition.
This procedure ensures that perturbations are applied incrementally and only to the extent necessary for successful attack generation.

As for the detailed attack process, we primarily employ synonym substitution to perturb words during the experiments. Specifically, as shown in Fig.~\ref{Perturbation_Strategy}, given a target word \( w_i \), we employ a synonym generation tool $f_{\text{tool}}$ to retrieve a set of \( r \) candidate synonyms, denoted as \( \tilde{w}_i^1, \tilde{w}_i^2, \ldots, \tilde{w}_i^r \). Among these candidates, we select the synonym that results in the greatest increase in the predicted probability of the farthest (i.e., least likely) label and use it to replace \( w_i \).
Formally, the selected synonym is given by:
\begin{equation}
\tilde{w}_i^{*} = \arg\max_{\tilde{w}_i^j \in f_{\text{tool}}(w_i)} \left[ P_{f_s}(y^{\text{far}} \mid \boldsymbol{x}_{\text{replaced}}^{(i,j)}) \right],
\end{equation}
where \( \boldsymbol{x}_{\text{replaced}}^{(i,j)} \) denotes the input text with \( w_i \) replaced by \( \tilde{w}_i^j \), and \( P_{f_s}(\cdot) \) denotes the prediction probability under the surrogate model.

\section{Experiment}
\subsection{Label Intersection Attack Success Rate Definition}\label{setloss}
Attack Success Rate (ASR) is a common metric for evaluating adversarial attacks for single-label classification but is inadequate for multi-label classification. 
To address this limitation, we propose the \textbf{Label Intersection Attack Success Rate (LI-ASR)}, designed to assess attack performance in such settings. 
For each text $\boldsymbol{x}_i^{\text{vic}}$ and its adversarial example $\tilde{\boldsymbol{x}}_i^{\text{vic}}$, we obtain the original labels $\boldsymbol{y}^{\text{ori}}_i$ and adversarial labels $\boldsymbol{y}^{\text{adv}}_i$ from the victim model. 
The attack effectiveness is defined as
$
1 - \frac{\left| \boldsymbol{y}_i^{\text{ori}} \cap \boldsymbol{y}_i^{\text{adv}} \right|}{\left| \boldsymbol{y}_i^{\text{ori}} \right|},
$
representing the proportion of original labels removed in the adversarial output. 
The overall LI-ASR is the mean of this value across all $n$ victim texts:
\begin{equation}
\text{LI-ASR} = 1 - \frac{1}{n} \sum_{i=1}^n \frac{\left| \boldsymbol{y}_i^{\text{ori}} \cap \boldsymbol{y}_i^{\text{adv}} \right|}{\left| \boldsymbol{y}_i^{\text{ori}} \right|},
\end{equation}
where $\left| \cdot \right|$ denotes set cardinality and $n$ represents the number of victim texts.


\subsection{Experimental Setup}

\textbf{Surrogate Model:} 
The surrogate model is a transformer-based architecture, which consists of $12$ hidden layers, each with a size of $768$. ``GELU''~\cite{lee2023gelu} is used as an activation function, with a dropout rate of $0.1$. The AdamW~\cite{zhou2024towards} optimizer is used for training, with a batch size of $32$ and a learning rate of $5e-5$ over $3$ epochs.


\textbf{Evaluation Metrics:} 
We employ several metrics to evaluate the effectiveness of our method. 
\ding{182} \textbf{Attack Success Rate (ASR):} ASR quantifies the proportion of successful attacks that deceive the victim model. A higher ASR indicates a more effective attack method.
\ding{183} \textbf{Number of Queries (Queries):} This metric reflects the total number of queries made to the victim model. A lower number of queries suggests a more efficient attack.
\ding{184} \textbf{Semantic Similarity (Similarity):} Semantic Similarity measures the average similarity between adversarial examples and victim texts. Following~\cite{wang2025no}, we use the T5 pre-trained model to vectorize both the original and adversarial texts, and compute the cosine similarity between the resulting vectors as the metric.
Higher similarity indicates a more successful attack strategy.
\ding{185} \textbf{LI-ASR:} Details regarding LI-ASR can be found in Section \ref{setloss}. A higher LI-ASR  indicates a more successful attack strategy.

\textbf{Dataset:}
We conduct experiments on six victim datasets. 
Specifically, in the main experiment of TDOA, we conduct experiments on two datasets: \textbf{SST5}~\cite{hsu2017hybrid} and \textbf{Emotion}~\cite{saravia-etal-2018-carer}.
The \textbf{Emotion} dataset, encompassing six emotions, is compiled from Twitter messages. The \textbf{SST5} dataset consists of five sentiment categories and is derived from movie reviews. In the additional experiment, we incorporate the TREC50 and TweetEval datasets for the classification task, as well as the Opus100 En–Zh and Opus100 En–Fr datasets for the translation task.
The detailed sources of the datasets are listed in Table~\ref{address}. 
\begin{table}[t]
\centering
\caption{Dataset URL}
\label{address}
\resizebox{1\columnwidth}{!}{%
\begin{tabular}{@{}c|c@{}}
\toprule
Dataset    & URL                                                                   \\ \midrule \midrule

Emotion    & \url{https://huggingface.co/datasets/dair-ai/emotion}                       \\
SST5       & \url{https://huggingface.co/datasets/SetFit/sst5}                           \\
TREC50     & \url{https://huggingface.co/datasets/CogComp/trec}                          \\ 
TweetEval  & \url{https://huggingface.co/datasets/cardiffnlp/tweet\_eval}                \\

Opus100 En–Zh     & \url{https://huggingface.co/datasets/Helsinki-NLP/opus-100/viewer/en-fr}                          \\ 
Opus100 En–Fr  & \url{https://huggingface.co/datasets/Helsinki-NLP/opus-100/viewer/en-zh}                \\

\bottomrule
\end{tabular}%
}
\end{table}

\textbf{Baselines:}
As no prior work has addressed attacks in the DO scenario, we adopt traditional textual attack methods. We focus on those applicable to the same setting as TDOA—namely, the hard-label black-box scenario where only the final predicted label is observable. Specifically, we evaluate CE Attack~\cite{formento2025confidence}, Emoji-attack~\cite{zhang2025emotiattack}, HQA~\cite{liu2024hqa}, Leap~\cite{ye2022leapattack},  and LimeAttack~\cite{zhu2024limeattack}. To enable a more comprehensive comparison of attack performance, we additionally include several white-box and soft-label black-box attack methods as baselines, including BAE~\cite{garg2020Bae}, CT-GAT~\cite{lv2023ct}, DWB~\cite{gao2018black}, FD~\cite{papernot2016crafting}, and TextBugger~\cite{ren2019generating}. 

\textbf{Victim Models:}
For the multi-label classification task, three publicly available multi-label models trained on the \textbf{Go-emotion} dataset are chosen as victim models. 
The backbone architectures of these victim models are BERT, DistilBERT, and RoBERTa, respectively.
For the LLM classification task, four LLMs are selected as the victim models.
The main experiments use GPT-4o and GPT-4.1 as victim models, while the additional experiments use Claude Sonnet 3.7 and DeepSeek-V3. Their URLs are listed in Table~\ref{url}, covering the GPT family (GPT-4o-mini, GPT-4o, GPT-4.1) and Claude Sonnet 3.7.

\begin{table}[t!]
\centering
\caption{Victim Model URL}
\label{url}
\resizebox{1\columnwidth}{!}{%
\begin{tabular}{@{}c|c@{}}
\toprule
Model & Url \\ \midrule \midrule
Distilbert & \url{https://huggingface.co/joeddav/distilbert-base-uncased-go-emotions-student} \\ 
BERT & \url{https://huggingface.co/bhadresh-savani/bert-base-go-emotion} \\ 
Roberta & \url{https://huggingface.co/bsingh/roberta_goEmotion} \\ 
GPT-4o               & \url{https://www.openai.com}                                       \\
GPT-4.1              & \url{https://www.openai.com}                                       \\
Claude Sonnet 3.7    & \url{https://www.anthropic.com}                                    \\
DeepSeek-V3          & \url{https://platform.deepseek.com}                                \\\bottomrule
\end{tabular}
}
\end{table}

 \textbf{Other Setups:} The clustering method used is K-means~\cite{ahmed2020k}, with a cluster number of $3$. The similarity threshold $\tau$ in Equation \ref{stop_equ} is $0.94$. We employ the T5 pre-trained model~\cite{ni2022sentence} as the embedding method.




\subsection{Main Results}
We conduct experiments in both the multi-label and LLMs scenarios. 
In the LLMs scenario, we utilize prompt learning~\cite{lei2024prompt} for classification tasks. For example, when using the SST5 dataset, we employ the following prompt: 
``\textit{Determine the label of the given text by selecting from the categories: very negative, negative, neutral, positive, or very positive.}''.
TDOA-1 and TDOA-5 denote the use of the top-1 and top-5 candidate adversarial examples generated by TDOA, respectively. Specifically, TDOA-1 relies on the top-1 candidate adversarial example and issues a single query, whereas TDOA-5 leverages the top-5 candidate adversarial examples and performs five queries.

\subsubsection{Multi-label Scenario}
We compare TDOA with other attack methods in Table \ref{other}. With 50 queries and access only to hard labels (no gradients or probabilities), TDOA achieves a maximum LI-ASR of 52.87\%. In addition, it consistently maintains a similarity score above 0.95 across all victim models and datasets, demonstrating strong textual consistency. Even when applying the semantically most distant label attack, TDOA does not markedly diminish text similarity. Meanwhile, the similarity threshold $\tau$ in Equation \ref{stop_equ} prevents excessive semantic drift.
Specifically, among the hard-label black-box attack methods (HQA, Leap, LimeAttack, and Emoji-Attack), 
DOTA achieves state-of-the-art (SOTA) performance on both LI-ASR and Queries metrics, surpassing the second-best method by an average of 16.81\% on LI-ASR. However, compared with white-box and soft-label black-box attacks, TDOA yields lower semantic similarity due to the stricter constraints of the hard-label black-box setting under which DOTA generates adversarial examples. Even so, TDOA consistently ranks second in semantic similarity among all hard-label black-box attack methods.
\begin{table*}[ht]
\centering
\caption{The results of TDOA and other attack methods.  ``Queries'' indicates the total number of queries, and ``Similarity'' indicates the average semantic similarity of original texts and adversarial texts. ``Bert'', ``DistilBERT'' and ``RoBERTa'' are the victim models. TDOA-1: uses the top-1 candidate adversarial example with one query.
TDOA-5: uses the top-5 candidate adversarial examples with five queries.}
\label{other}
\resizebox{1\textwidth}{!}{
\begin{tabular}{@{}cccccccccccccc@{}}
\toprule
\multirow{2}{*}{Datasets}                      & \multirow{2}{*}{Methods}          & \multirow{2}{*}{Gradient} & \multirow{2}{*}{Probability} & \multirow{2}{*}{Hard-label} & \multicolumn{3}{c}{Bert}                                            & \multicolumn{3}{c}{DistilBERT}                                      & \multicolumn{3}{c}{RoBERTa}                    \\ \cmidrule(l){6-14} 
                                               &                                   &                           &                              &                             & LI-ASR ↑          & Similarity ↑     & \multicolumn{1}{c|}{Queries ↓}    & LI-ASR ↑          & Similarity ↑    & \multicolumn{1}{c|}{Queries ↓}    & LI-ASR ↑          & Similarity ↑    & Queries ↓    \\ \midrule \midrule
\multicolumn{1}{c|}{\multirow{12}{*}{Emotion}} & \multicolumn{1}{c|}{Bae}          & \XSolidBrush                         & \CheckmarkBold                            & \multicolumn{1}{c|}{\CheckmarkBold}      & 29.34\%          & 0.929          & \multicolumn{1}{c|}{21.70}      & 29.04\%          & 0.927          & \multicolumn{1}{c|}{21.75}      & 33.20\%          & 0.923          & 21.82      \\
\multicolumn{1}{c|}{}                          & \multicolumn{1}{c|}{CT-GAT}       & \XSolidBrush                         & \CheckmarkBold                            & \multicolumn{1}{c|}{\CheckmarkBold}      & 16.64\%          & 0.958          & \multicolumn{1}{c|}{21.03}      & 16.83\%          & 0.969          & \multicolumn{1}{c|}{20.83}      & 9.31\%           & 0.989          & 20.86      \\
\multicolumn{1}{c|}{}                          & \multicolumn{1}{c|}{DWB}          & \XSolidBrush                         & \CheckmarkBold                            & \multicolumn{1}{c|}{\CheckmarkBold}      & 34.08\%          & 0.980          & \multicolumn{1}{c|}{21.14}      & 39.04\%          & 0.975          & \multicolumn{1}{c|}{21.01}      & 33.65\%          & 0.979          & 22.41      \\
\multicolumn{1}{c|}{}                          & \multicolumn{1}{c|}{FD}           & \CheckmarkBold                         & \CheckmarkBold                            & \multicolumn{1}{c|}{\CheckmarkBold}      & 28.73\%          & 0.936          & \multicolumn{1}{c|}{11.07}      & 24.15\%          & 0.940          & \multicolumn{1}{c|}{10.82}      & 8.06\%           & 0.981          & 11.57      \\
\multicolumn{1}{c|}{}                          & \multicolumn{1}{c|}{TextBugger}   & \CheckmarkBold                         & \CheckmarkBold                            & \multicolumn{1}{c|}{\CheckmarkBold}      & 32.05\%          & 0.981          & \multicolumn{1}{c|}{28.30}      & 33.57\%          & 0.980          & \multicolumn{1}{c|}{28.74}      & 32.42\%          & 0.977          & 30.33      \\ \cmidrule(l){2-14} 
\multicolumn{1}{c|}{}                          & \multicolumn{1}{c|}{CE Attack}    & \XSolidBrush                         & \XSolidBrush                            & \multicolumn{1}{c|}{\CheckmarkBold}      & 23.37\%          & 0.972          & \multicolumn{1}{c|}{9.29}       & 17.35\%          & 0.961          & \multicolumn{1}{c|}{9.52}       & 26.67\%          & 0.958          & 10.33      \\
\multicolumn{1}{c|}{}                          & \multicolumn{1}{c|}{Emoji Attack} & \XSolidBrush                         & \XSolidBrush                            & \multicolumn{1}{c|}{\CheckmarkBold}      & 10.39\%          & 0.973          & \multicolumn{1}{c|}{10.14}      & 14.63\%          & {\underline{0.968} }    & \multicolumn{1}{c|}{12.52}      & 24.67\%          & 0.959          & 12.05      \\
\multicolumn{1}{c|}{}                          & \multicolumn{1}{c|}{HQA}          & \XSolidBrush                         & \XSolidBrush                            & \multicolumn{1}{c|}{\CheckmarkBold}      & 22.15\%          & 0.969          & \multicolumn{1}{c|}{10.08}      & 18.26\%          & 0.964          & \multicolumn{1}{c|}{9.97}       & 28.16\%          & 0.949          & 14.40      \\
\multicolumn{1}{c|}{}                          & \multicolumn{1}{c|}{Leap}         & \XSolidBrush                         & \XSolidBrush                            & \multicolumn{1}{c|}{\CheckmarkBold}      & 24.88\%          & 0.965          & \multicolumn{1}{c|}{10.15}      & 24.91\%          & 0.958          & \multicolumn{1}{c|}{9.96}       & 25.92\%          & 0.952          & 12.19      \\
\multicolumn{1}{c|}{}                          & \multicolumn{1}{c|}{LimeAttack}   & \XSolidBrush                         & \XSolidBrush                            & \multicolumn{1}{c|}{\CheckmarkBold}      & 13.81\%          & \textbf{0.979} & \multicolumn{1}{c|}{14.91}      & 10.42\%          & \textbf{0.978} & \multicolumn{1}{c|}{12.14}      & 15.88\%          & \textbf{0.983} & 16.04      \\ \cmidrule(l){2-14} 
\multicolumn{1}{c|}{}                          & \multicolumn{1}{c|}{TDOA-1}       & \XSolidBrush                         & \XSolidBrush                            & \multicolumn{1}{c|}{\CheckmarkBold}      & {\underline{42.32\%}}    & {\underline{0.978}}    & \multicolumn{1}{c|}{\textbf{1}} & {\underline{52.87\%}}    & 0.961          & \multicolumn{1}{c|}{\textbf{1}} & {\underline{41.55\%}}    & {\underline{0.962}}    & \textbf{1}    \\
\multicolumn{1}{c|}{}                          & \multicolumn{1}{c|}{TDOA-5}       & \XSolidBrush                         & \XSolidBrush                            & \multicolumn{1}{c|}{\CheckmarkBold}      & \textbf{67.30\%} & 0.977          & \multicolumn{1}{c|}{\underline{5}}    & \textbf{76.31\%} & 0.961          & \multicolumn{1}{c|}{\underline{5}}    & \textbf{66.24\%} & 0.960          & \underline{5} \\ \midrule
\multicolumn{1}{c|}{\multirow{12}{*}{SST5}}    & \multicolumn{1}{c|}{Bae}          & \XSolidBrush                         & \CheckmarkBold                            & \multicolumn{1}{c|}{\CheckmarkBold}      & 28.34\%          & 0.886          & \multicolumn{1}{c|}{21.47}      & 34.85\%          & 0.887          & \multicolumn{1}{c|}{21.36}      & 28.36\%          & 0.888          & 21.45      \\
\multicolumn{1}{c|}{}                          & \multicolumn{1}{c|}{CT-GAT}       & \XSolidBrush                         & \CheckmarkBold                            & \multicolumn{1}{c|}{\CheckmarkBold}      & 14.57\%          & 0.964          & \multicolumn{1}{c|}{22.50}      & 16.91\%          & 0.971          & \multicolumn{1}{c|}{21.99}      & 15.49\%          & 0.967          & 22.35      \\
\multicolumn{1}{c|}{}                          & \multicolumn{1}{c|}{DWB}          & \XSolidBrush                         & \CheckmarkBold                            & \multicolumn{1}{c|}{\CheckmarkBold}      & 20.92\%          & 0.979          & \multicolumn{1}{c|}{20.73}      & 37.56\%          & 0.973          & \multicolumn{1}{c|}{19.03}      & 24.75\%          & 0.975          & 22.58      \\
\multicolumn{1}{c|}{}                          & \multicolumn{1}{c|}{FD}           & \CheckmarkBold                         & \CheckmarkBold                            & \multicolumn{1}{c|}{\CheckmarkBold}      & 23.08\%          & 0.939          & \multicolumn{1}{c|}{12.55}      & 20.65\%          & 0.949          & \multicolumn{1}{c|}{10.58}      & 5.37\%           & 0.983          & 10.03      \\
\multicolumn{1}{c|}{}                          & \multicolumn{1}{c|}{TextBugger}   & \CheckmarkBold                         & \CheckmarkBold                            & \multicolumn{1}{c|}{\CheckmarkBold}      & 22.20\%          & 0.979          & \multicolumn{1}{c|}{26.68}      & 28.25\%          & 0.976          & \multicolumn{1}{c|}{21.28}      & 21.52\%          & 0.977          & 28.48      \\ \cmidrule(l){2-14} 
\multicolumn{1}{c|}{}                          & \multicolumn{1}{c|}{CE-Attack}    & \XSolidBrush                         & \XSolidBrush                            & \multicolumn{1}{c|}{\CheckmarkBold}      & 14.92\%          & 0.968          & \multicolumn{1}{c|}{12.73}      & 16.93\%          & 0.959          & \multicolumn{1}{c|}{11.50}      & 18.29\%          & 0.958          & 12.79      \\
\multicolumn{1}{c|}{}                          & \multicolumn{1}{c|}{Emoji Attack} & \XSolidBrush                         & \XSolidBrush                            & \multicolumn{1}{c|}{\CheckmarkBold}      & 14.70\%          & 0.962          & \multicolumn{1}{c|}{12.18}      & 14.69\%          & 0.961          & \multicolumn{1}{c|}{13.25}      & 20.82\%          & 0.959          & 10.83      \\
\multicolumn{1}{c|}{}                          & \multicolumn{1}{c|}{HQA}          & \XSolidBrush                         & \XSolidBrush                            & \multicolumn{1}{c|}{\CheckmarkBold}      & 19.35\%          & 0.953          & \multicolumn{1}{c|}{12.71}      & 20.25\%          & 0.960          & \multicolumn{1}{c|}{8.88}       & 22.24\%          & 0.950          & 13.46      \\
\multicolumn{1}{c|}{}                          & \multicolumn{1}{c|}{Leap}         & \XSolidBrush                         & \XSolidBrush                            & \multicolumn{1}{c|}{\CheckmarkBold}      & 19.82\%          & 0.958          & \multicolumn{1}{c|}{12.57}      & 24.45\%          & 0.958          & \multicolumn{1}{c|}{9.50}       & 21.20\%          & 0.955          & 12.67      \\
\multicolumn{1}{c|}{}                          & \multicolumn{1}{c|}{LimeAttack}   & \XSolidBrush                         & \XSolidBrush                            & \multicolumn{1}{c|}{\CheckmarkBold}      & 13.62\%          & \textbf{0.983} & \multicolumn{1}{c|}{16.35}      & 16.01\%          & \textbf{0.976} & \multicolumn{1}{c|}{21.85}      & 15.43\%          & \textbf{0.982} & 22.55      \\ \cmidrule(l){2-14} 
\multicolumn{1}{c|}{}                          & \multicolumn{1}{c|}{TDOA-1}       & \XSolidBrush                         & \XSolidBrush                            & \multicolumn{1}{c|}{\CheckmarkBold}      & 32.11\%          & 0.969          & \multicolumn{1}{c|}{\textbf{1}} & {\underline{46.09\%}}    & 0.961          & \multicolumn{1}{c|}{\textbf{1}} & {\underline{29.34\%}}    & {\underline{0.965}}    & \textbf{1} \\
\multicolumn{1}{c|}{}                          & \multicolumn{1}{c|}{TDOA-5}                            & \XSolidBrush                         & \XSolidBrush                            & \multicolumn{1}{c|}{\CheckmarkBold}                          & \textbf{50.77\%} & {\underline{0.961}}    &\multicolumn{1}{c|} {\underline{5}}                         & \textbf{69.65\%} & {\underline{0.962}}    & \multicolumn{1}{c|}{\underline{5}}                         & \textbf{43.32\%} & 0.964          & {\underline{5} }    \\ \bottomrule
\end{tabular}
}
\end{table*}

\subsubsection{LLMs Scenario}
Since GPT-4.1 and GPT-4o only provide final predictions without gradients, losses, or class probabilities, this task is categorized as a hard-label black-box attack. We therefore compare TDOA against representative baselines—CE Attack, Emoji-attack, HQA, Leap, and LimeAttack.
Table \ref{llm} reports the experimental results for GPT-4.1~\cite{ho2025analisis} and GPT-4o~\cite{hurst2024gpt} LLMs. 
TDOA refers to a setting in which only a single candidate adversarial example is evaluated, whereas TDOA-5 allows for up to five candidates, with the attack deemed successful if any one of them successfully attacks the victim model.
Experimental results show that TDOA attains an average ASR of 45.53\% on Emotion and 41.41\% on SST-5, outperforming other methods restricted to fewer than 30 queries per sample but lagging behind those with unrestricted querying. By contrast, TDOA-5 achieves 69.43\% on Emotion and 69.37\% on SST-5, surpassing even baselines with unlimited queries.

\begin{table}[t]
\caption{The result of TDOA and other attack methods for GPT-4o and GPT-4.1.}  
\label{llm}
\resizebox{0.485\textwidth}{!}{%
\begin{tabular}{@{}cccccccc@{}}
\toprule
\multirow{2}{*}{Datasets}                     & \multirow{2}{*}{Methods}          & \multicolumn{3}{c}{GPT-4o}                                               & \multicolumn{3}{c}{GPT-4.1}                         \\ \cmidrule(l){3-8} 
                                              &                                   & ASR ↑             & Similarity ↑ & \multicolumn{1}{c|}{Queries ↓}    & ASR ↑             & Similarity ↑ & Queries ↓   \\ \midrule \midrule
\multicolumn{1}{c|}{\multirow{7}{*}{Emotion}} & \multicolumn{1}{c|}{CE Attack}    & {\underline{58.0\%}}    & 0.931               & \multicolumn{1}{c|}{16.92}      & {\underline{60.3\%}}    & 0.926               & 15.15      \\
\multicolumn{1}{c|}{}                         & \multicolumn{1}{c|}{Emoji Attack} & 49.9\%          & 0.943               & \multicolumn{1}{c|}{8.80}       & 46.0\%          & 0.945               & 9.50       \\
\multicolumn{1}{c|}{}                         & \multicolumn{1}{c|}{HQA}          & 32.4\%          & 0.945               & \multicolumn{1}{c|}{6.82}       & 32.4\%          & 0.947               & 6.86       \\
\multicolumn{1}{c|}{}                         & \multicolumn{1}{c|}{Leap}         & 30.3\%          & 0.944               & \multicolumn{1}{c|}{6.67}       & 29.3\%          & 0.946               & 6.60       \\
\multicolumn{1}{c|}{}                         & \multicolumn{1}{c|}{LimeAttack}   & 23.1\%          & \textbf{0.981}      & \multicolumn{1}{c|}{26.03}      & 24.9\%          & \textbf{0.990}      & 25.98      \\ \cmidrule(l){2-8} 
\multicolumn{1}{c|}{}                         & \multicolumn{1}{c|}{TDOA-1}       & 40.8\%          & {\underline{0.973}}         & \multicolumn{1}{c|}{\textbf{1}} & 50.3\%          & {\underline{0.974}}         & \textbf{1} \\
\multicolumn{1}{c|}{}                         & \multicolumn{1}{c|}{TDOA-5}       & \textbf{63.7\%} & 0.965               & \multicolumn{1}{c|}{5}          & \textbf{75.2\%} & 0.967               & 5          \\ \midrule
\multicolumn{1}{c|}{\multirow{7}{*}{SST5}}    & \multicolumn{1}{c|}{CE Attack}    & {\underline{55.6\%}}    & 0.938               & \multicolumn{1}{c|}{12.10}      & {\underline{66.8\%}}    & 0.943               & 16.22      \\
\multicolumn{1}{c|}{}                         & \multicolumn{1}{c|}{Emoji Attack} & 49.6\%          & 0.972               & \multicolumn{1}{c|}{9.65}       & 45.3\%          & 0.966               & 9.07       \\
\multicolumn{1}{c|}{}                         & \multicolumn{1}{c|}{HQA}          & 36.3\%          & 0.953               & \multicolumn{1}{c|}{6.90}       & 32.4\%          & 0.947               & 6.86       \\
\multicolumn{1}{c|}{}                         & \multicolumn{1}{c|}{Leap}         & 36.2\%          & 0.951               & \multicolumn{1}{c|}{7.08}       & 29.3\%          & 0.946               & 6.60       \\
\multicolumn{1}{c|}{}                         & \multicolumn{1}{c|}{LimeAttack}   & 31.1\%          & \textbf{0.988}      & \multicolumn{1}{c|}{25.48}      & 30.5\%          & \textbf{0.986}      & 25.62      \\ \cmidrule(l){2-8} 
\multicolumn{1}{c|}{}                         & \multicolumn{1}{c|}{TDOA-1}       & 32.0\%          & {\underline{0.973}}         & \multicolumn{1}{c|}{\textbf{1}} & 50.8\%          & 0.972               & \textbf{1} \\
\multicolumn{1}{c|}{}                         & \multicolumn{1}{c|}{TDOA-5}                            & \textbf{56.1\%} & {\underline{0.967}}         & \multicolumn{1}{c|}{\underline{5}}                         & \textbf{80.7\%} & {\underline{0.968}}         & 5          \\ \bottomrule
\end{tabular}
}
\end{table}

\subsection{Ablation Study}
\begin{figure}[ht]
    \centering
        \includegraphics[width=0.5\textwidth]{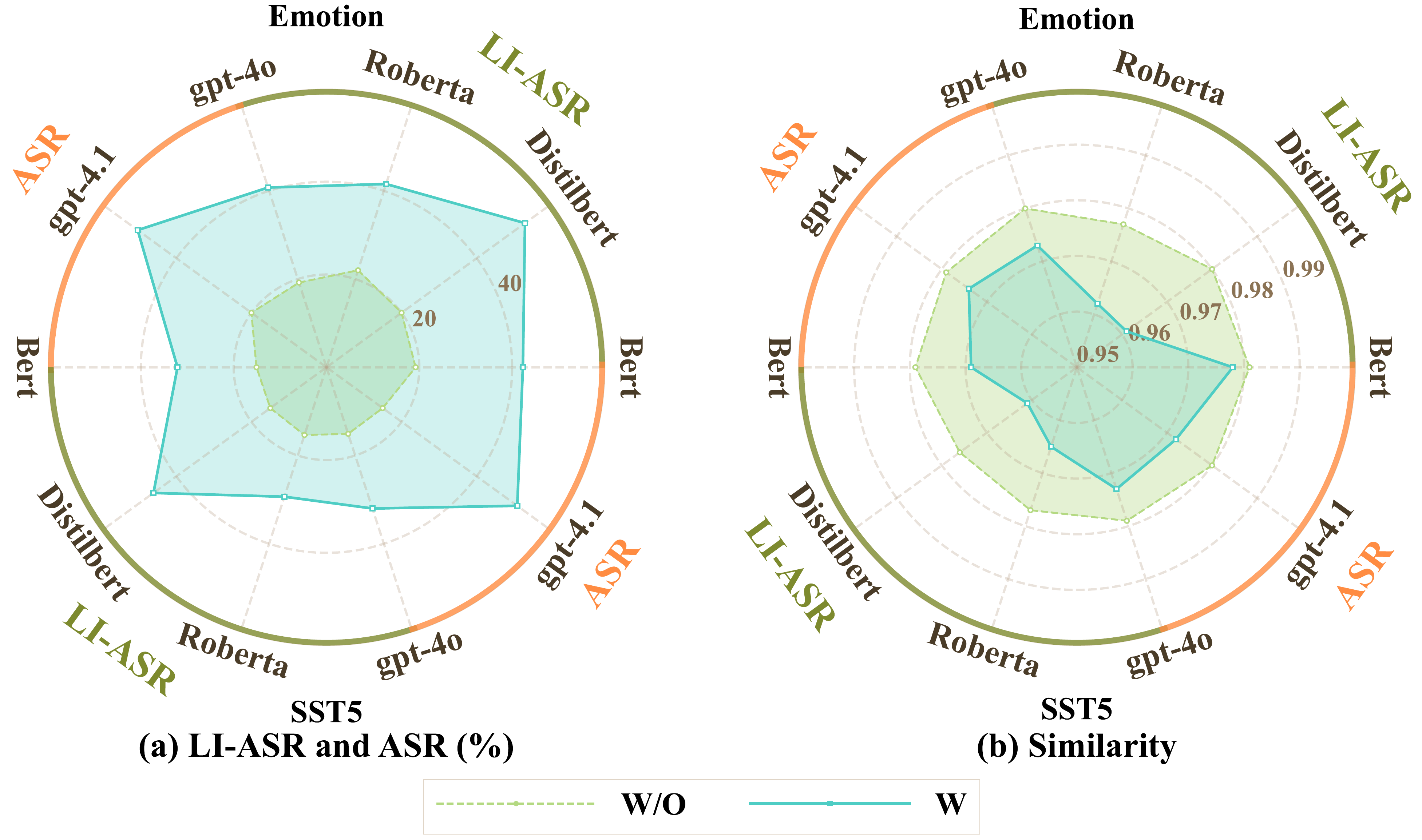}

\caption{LI-ASR(\%), ASR(\%) and Similarity of with and without the clustering-based surrogate model}\label{clustering-based}
\end{figure}

\textbf{The impact of the clustering-based surrogate model.} 
To demonstrate the importance of the clustering-based surrogate model, we evaluate TDOA attacks using traditional surrogate models that closely resemble the victim model. In the multi-label scenario, since the victim model is trained on the Go-Emotions dataset, we also select other multi-label models trained on the same dataset as surrogates; the address of these models is provided in Table \ref{url}, referred to as Model A. In the LLM scenario, since the training data for ChatGPT-4 and ChatGPT-4.1 is unavailable, we employ Claude LLM as the surrogate model. As illustrated in Figure \ref{clustering-based}, excluding the clustering-based surrogate model substantially degrades TDOA’s performance, with average decreases of 26.15\% in LI-ASR and 22.95\% in accuracy, despite a modest average increase of 0.01 in similarity. We argue that this trade-off—sacrificing 0.01 similarity for a significant improvement in attack effectiveness—is well justified.

\begin{figure}[ht]
    \centering
        \includegraphics[width=0.5\textwidth]{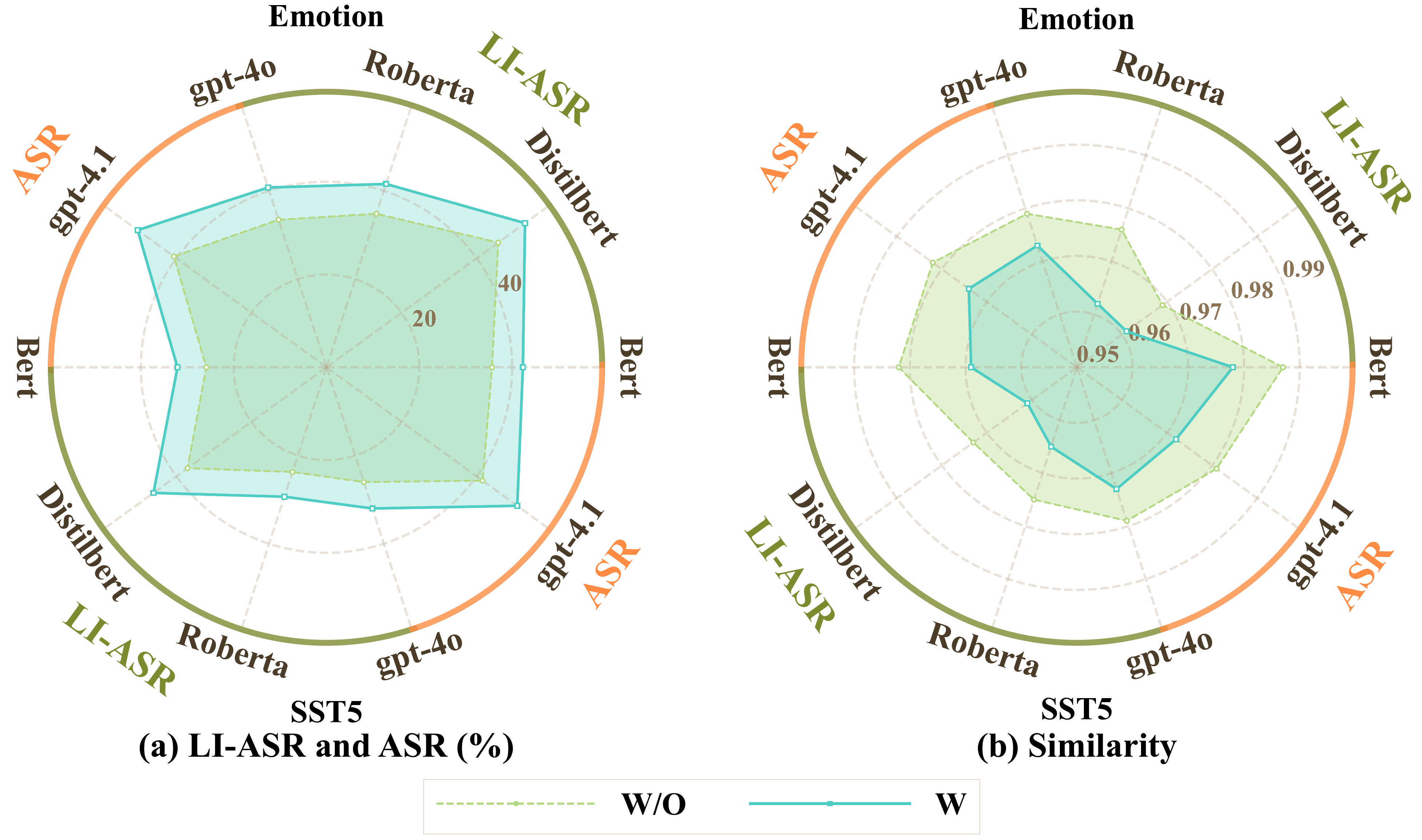}

\caption{ LI-ASR(\%), ASR(\%) and Similarity of with and without the farthest  label  target attack}\label{farthest_distance_label_attack}
\end{figure}

\textbf{The impact of the farthest label target attack:}  
Figure \ref{farthest_distance_label_attack}
shows the results with and without applying the farthest label target attack. When the farthest-label targeted attack is applied, both the ASR and LI-ASR of TODA increase, while the similarity score decreases. Specifically, the average ASR decreases from 45.53\% to 37.05\%, and the average LI-ASR rises  from 40.71\% to 33.81\%, whereas the average similarity declines from 0.978 to 0.969. To achieve better attack performance, we employ the farthest label targeted attack in TODA.

\begin{figure}[ht]
    \centering
        \includegraphics[width=0.485\textwidth]{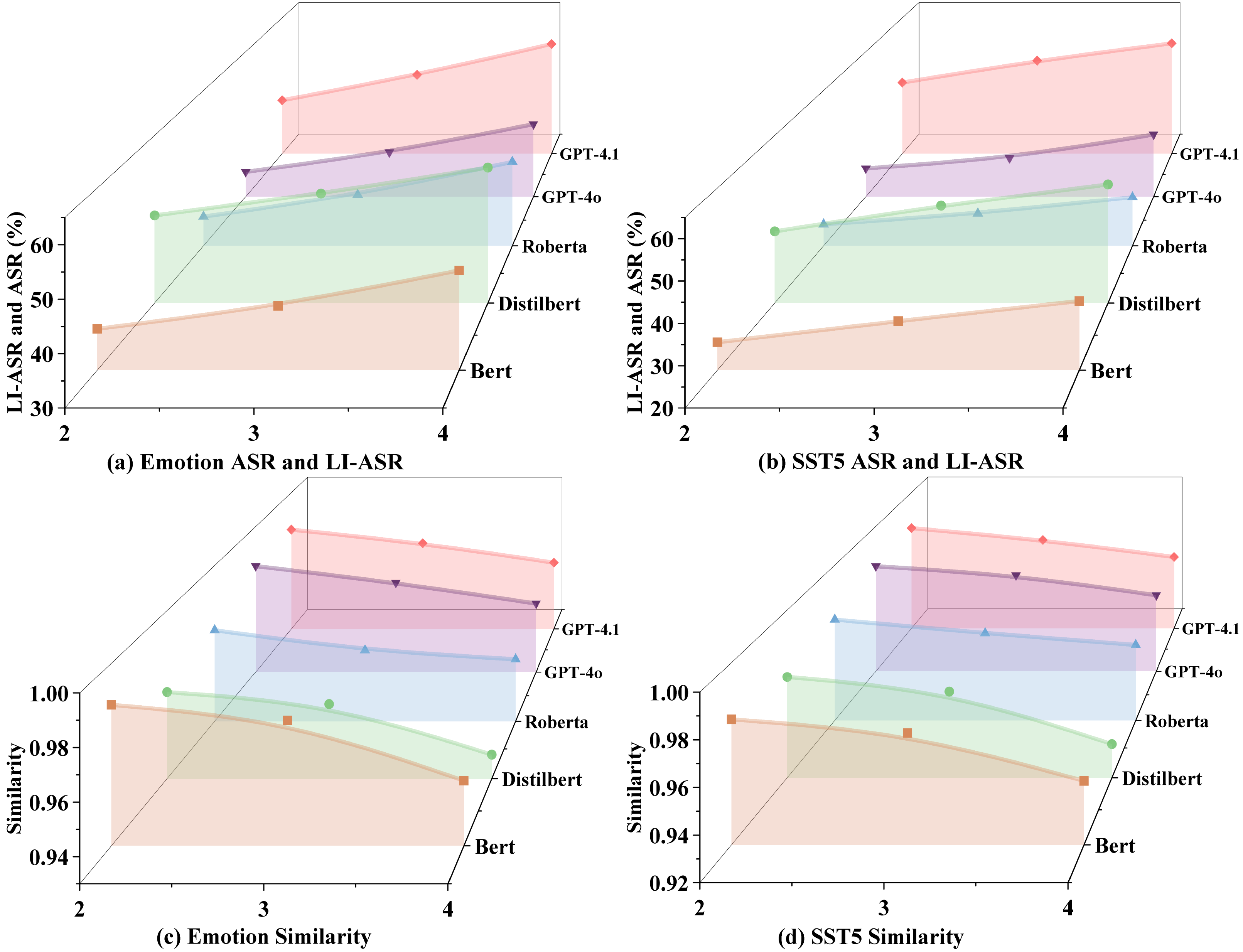}

\caption{LI-ASR(\%), ASR(\%) and Similarity of different cluster numbers}\label{CN1}
\end{figure}

\textbf{The impact of different cluster numbers:}
Figure \ref{CN1} shows that as the number of clusters increases from 2 to 3 and then to 4, the average LI-ASR rises from 35.84
\% to 40.71\% and 46.56\%, and ASR rises from 38.03
\% to 43.47\% and 50.40\%. In contrast, similarity drops from 0.976 to 0.970 and 0.947.
This trend is attributed to the farthest label targeted attack, which is integrated into TDOA. With a larger number of clusters, the farthest label becomes more semantically distant from the original text, producing adversarial examples that deviate further from the original semantics. As a result, similarity declines, whereas LI-ASR and ASR increase.
To balance attack effectiveness and semantic preservation, we set the number of clusters in TDOA to $3$, based on LI-ASR, ASR, and similarity.

\textbf{The impact of different vectorization methods:}
\begin{figure}[ht]
    \centering
        \includegraphics[width=0.485\textwidth]{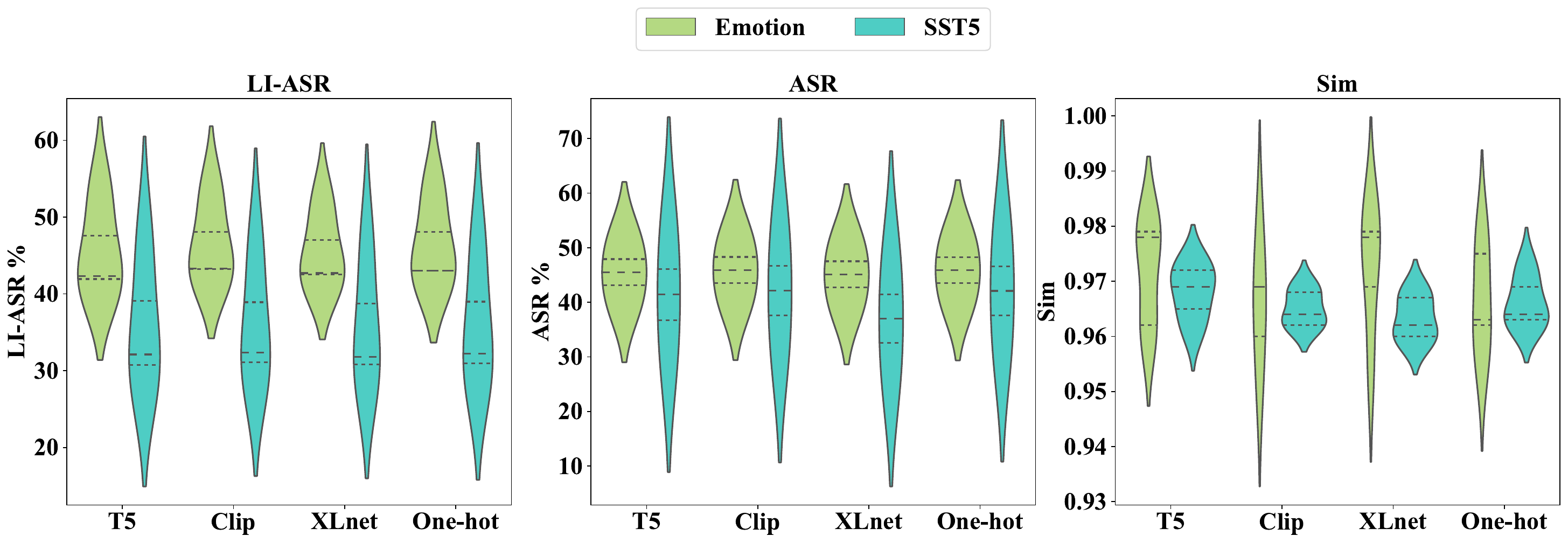}

\caption{LI-ASR(\%), ASR(\%), and Similarity of different vectorization methods}\label{VM1}
\end{figure}
Figure \ref{VM1} presents the attack results for different vectorization methods (e.g., T5~\cite{ni2022sentence}, CLIP~\cite{radford2021learning}, XLNet~\cite{yang2019xlnet}, One-hot~\cite{okada2019efficient}) under the same query and attack perturbation conditions. 
The results show minimal impact of vectorization methods on attack performance, with LI-ASR and ASR varying between 40–50\%, and no method consistently outperforming others, suggesting the influence of different vectorization methods on attack efficacy appears random.

\begin{figure}[ht]
    \centering
        \includegraphics[width=0.485\textwidth]{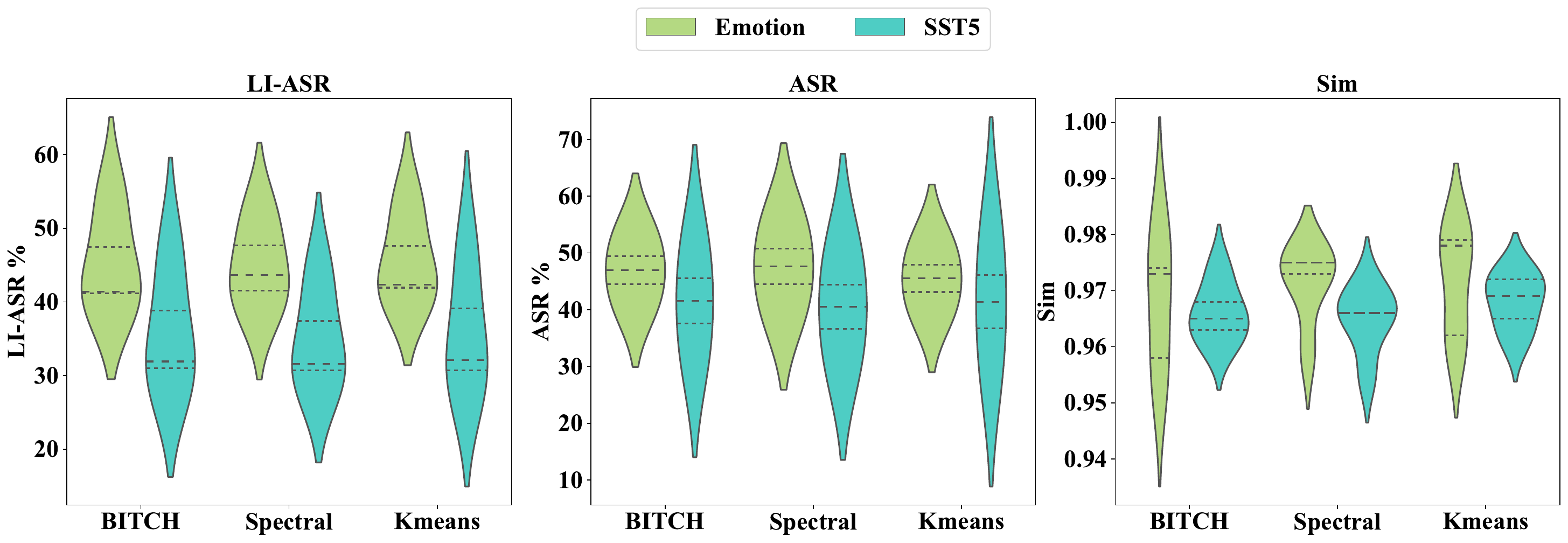}

\caption{LI-ASR(\%), ASR(\%) and Similarity of different cluster methods}\label{cluster_method}
\end{figure}

\textbf{The impact of different cluster methods:}
Figure \ref{cluster_method} presents the results of different clustering methods (K-means~\cite{ahmed2020k}, BIRCH~\cite{zhang1996birch} and Spectral Clustering~\cite{von2007tutorial}). Our findings suggest that clustering methods introduce random variability in TDOA's attack performance.
For example, in the Emotion dataset, the LI-ASR, ASR, and similarity scores across the three clustering methods vary between 41.36\%–53.65\%, 50.30\%–51.90\%, and 0.954-0.992, respectively. None of the clustering methods consistently achieves SOTA performance.

In summary, vectorization and clustering methods introduce stochastic variations in TODA's attack effectiveness. Moreover, increasing the number of clusters enhances both LI-ASR and ASR, but diminishes semantic similarity. Additionally, employing the farthest label targeted attack further improves attack performance.
\section{More Experiment}
\subsection{Scalability to Other Task}
We observe that machine translation can be viewed as a special case of classification, where each ground-truth translation corresponds to a label. Unlike traditional classification tasks, the label space for translation is infinite. Therefore, we consider translation tasks as a special form of a dynamic output scenario. In our experiments, we focus on English-to-Chinese (En-Zh) and English-to-French (En-Fr) translation tasks, using \textit{Baidu Translate}~\cite{baidu2019translate}, \textit{Ali Translate}~\cite{alibaba2020translate}, and \textit{Google Translate} as victim models. We use the Opus100 En-Zh and Opus100 En-Fr datasets~\cite{huang2010transcription} as victim datasets. Meanwhile, we adopt the hard-label black-box translation attack methods PROTES~\cite{chertkov2023translate}, TransFool~\cite{sadrizadehtransfool}, and Morpheus~\cite{tan-etal-2020-morphin} as baselines and use Relative Decrease in BLEU (RDBLEU)~\cite{papineni2002bleu}, Relative Decrease in chrF (RDchrF)~\cite{popovic2015chrf}, and semantic similarity as evaluation metrics. Table \ref{Trans} presents the attack results of TDOA in the translation task. TDOA achieves SOTA performance on translation tasks based on the RDBLEU and RDchrF metrics, while also yielding substantial improvements in similarity scores. Specifically, TDOA reaches up to 0.64 RDBLEU, 0.62 RDchrF, and a similarity score of 0.881.

\begin{table*}[h]
\caption{The result of TDOA and other attack methods in GPT-4o and GPT-4.1.  ``Queries'' indicates the total number of queries, and ``Similarity'' indicates the average semantic similarity of original texts and adversarial texts. ``DistilBERT'' and ``RoBERTa'' are the victim models.}
\label{Trans}
\resizebox{1\textwidth}{!}{%
\begin{tabular}{@{}cccccccccccccc@{}}
\toprule
\multirow{2}{*}{Datasets}                           & \multirow{2}{*}{Method}        & \multicolumn{4}{c}{Google Translate}                                             & \multicolumn{4}{c}{Ali Translate}                                                & \multicolumn{4}{c}{Baidu Translate}                         \\ \cmidrule(l){3-14} 
                                                    &                                & RDBLEU↑       & RDchrF ↑      & Similarity ↑   & \multicolumn{1}{c|}{Queries ↓}  & RDBLEU ↑      & RDchrF ↑      & Similarity ↑   & \multicolumn{1}{c|}{Queries ↓}  & RDBLEU ↑      & RDchrF ↑      & Similarity ↑   & Queries ↓  \\ \midrule \midrule
\multicolumn{1}{c|}{\multirow{5}{*}{Opus100 En-Fr}} & \multicolumn{1}{c|}{PROTES}    & 0.20          & 0.19          & 0.694          & \multicolumn{1}{c|}{45.71}      & 0.38          & 0.38          & 0.677          & \multicolumn{1}{c|}{34.81}      & 0.36          & 0.37          & 0.678          & 40.76      \\
\multicolumn{1}{c|}{}                               & \multicolumn{1}{c|}{TransFool} & 0.21          & 0.21          & 0.834          & \multicolumn{1}{c|}{18.23}      & 0.24          & 0.25          & 0.812          & \multicolumn{1}{c|}{11.27}      & 0.23          & 0.23          & 0.812          & 12.45      \\
\multicolumn{1}{c|}{}                               & \multicolumn{1}{c|}{Morpheus}  & 0.15          & 0.16          & \textbf{0.891} & \multicolumn{1}{c|}{5.64}       & 0.14          & 0.15          & \textbf{0.881} & \multicolumn{1}{c|}{5.11}       & 0.13          & 0.13          & 0.871          & 4.87       \\ \cmidrule(l){2-14} 
\multicolumn{1}{c|}{}                               & \multicolumn{1}{c|}{TODA-1}    & {\underline{0.23}}    & {\underline{0.22}}    & {\underline{0.878}}    & \multicolumn{1}{c|}{\textbf{1}} & {\underline{0.41}}    & {\underline{0.42}}    & \textbf{0.881} & \multicolumn{1}{c|}{\textbf{1}} & {\underline{0.40}}    & {\underline{0.40}}    & \textbf{0.879} & \textbf{1} \\
\multicolumn{1}{c|}{}                               & \multicolumn{1}{c|}{TDOA-5}    & \textbf{0.27} & \textbf{0.28} & 0.871          & \multicolumn{1}{c|}{\underline{5}}    & \textbf{0.49} & \textbf{0.49} & {\underline{0.878}}    & \multicolumn{1}{c|}{\underline{5}}    & \textbf{0.46} & \textbf{0.45} & {\underline{0.876}}    & \underline{5}    \\ \midrule
\multicolumn{1}{c|}{\multirow{5}{*}{Opus100 En-Zh}} & \multicolumn{1}{c|}{PROTES}    & 0.38          & 0.37          & 0.711          & \multicolumn{1}{c|}{57.23}      & 0.64          & 0.63          & 0.669          & \multicolumn{1}{c|}{39.46}      & \textbf{0.68} & \textbf{0.68} & 0.669          & 46.36      \\
\multicolumn{1}{c|}{}                               & \multicolumn{1}{c|}{TransFool} & 0.41          & 0.41          & 0.842          & \multicolumn{1}{c|}{11.54}      & 0.57          & 0.57          & 0.822          & \multicolumn{1}{c|}{8.30}       & 0.54          & 0.56          & 0.822          & 7.57       \\
\multicolumn{1}{c|}{}                               & \multicolumn{1}{c|}{Morpheus}  & 0.30          & 0.30          & 0.821          & \multicolumn{1}{c|}{5.67}       & 0.45          & 0.48          & 0.831          & \multicolumn{1}{c|}{4.33}       & 0.42          & 0.42          & 0.841          & 4.47       \\ \cmidrule(l){2-14} 
\multicolumn{1}{c|}{}                               & \multicolumn{1}{c|}{TODA-1}    & {\underline{0.41}}    & {\underline{0.41}}    & \textbf{0.868} & \multicolumn{1}{c|}{\textbf{1}} & {\underline{0.57}}    & {\underline{0.56}}    & \textbf{0.866} & \multicolumn{1}{c|}{\textbf{1}} & 0.57          & 0.57          & \textbf{0.873} & \textbf{1} \\
\multicolumn{1}{c|}{}                               & \multicolumn{1}{c|}{TDOA-5}    & \textbf{0.47} & \textbf{0.48} & {\underline{0.863}}    & \multicolumn{1}{c|}{\underline{5}}    & \textbf{0.64} & \textbf{0.62} & {\underline{0.863}}    & \multicolumn{1}{c|}{\underline{5}}    & {\underline{0.60}}    & {\underline{0.60}}    & {\underline{0.870}}    & \underline{5}    \\ \bottomrule
\end{tabular}
}
\end{table*}

\subsection{Results in the Static Output Space}
Although TDOA is originally designed for generating adversarial examples in dynamic-output scenarios, it can also achieve SOTA results in traditional static-output settings. As shown in Table~\ref{NO-DO}, we conduct experiments on static-output victim models DistilBERT-SST5 and RoBERTa-SST5, where TDOA attains a highest ASR of up to 82.68\%.

\begin{table}[h]
\centering
\caption{
Experimental results in the static output space
}
\label{NO-DO}
\resizebox{0.48\textwidth}{!}{%
\begin{tabular}{@{}c|ccc|ccc@{}}
\toprule
Victim Models & \multicolumn{3}{c|}{DistilBERT-SST5}         & \multicolumn{3}{c}{RoBERTa-SST5}             \\ \midrule
Methods       & ASR ↑           & Similarity ↑    & Queries ↓    & ASR ↑          & Similarity ↑    & Queries ↓    \\ \midrule \midrule
Bae           & 52.5\%          & 0.855          & 24.27      & 55.5\%            & 0.855          & 25.08      \\
CT-GAT        & 42.3\%            & 0.958          & 26.30      & 37.31          & 0.947          & 38.60      \\
DWB           & 57.0\%           & 0.930          & 27.96      & 48.9\%            & 0.951          & 30.21      \\
FD            & 40.5\%            & 0.929          & 68.12      & 16.4\%            & 0.954          & 16.33      \\
TextBugger    & 79.1\%            & 0.952          & 40.26      & 69.1\%            & 0.951          & 49.13      \\ \midrule
CE Attack     & 42.8\%            & 0.960          & 46.47      & 31.6\%            & 0.959          & 58.07      \\
Emoji Attack  & 17.2\%            & 0.977          & 7.72       & 16.5\%            & 0.977          & 9.33       \\
HQA           & 37.5\%            & 0.948          & 34.15      & 28.5\%            & 0.944          & 35.19      \\
Leap          & 41.4\%            & 0.941          & 53.92      & 33.6\%          & 0.934          & 54.28      \\
LimeAttack    & 20.3\%            & \textbf{0.982} & 24.84      & 17.9\%            & \textbf{0.982} & 23.37      \\ \midrule
TDOA-1        & {\underline{50.6\%  }}    & 0.954          & \textbf{1} & {\underline{42.5\%  }}    & 0.952          & \textbf{1} \\
TDOA-5        & \textbf{82.68} & 0.949          & \underline{5}    & \textbf{76.8\%  } & 0.950          & \underline{5}    \\ \bottomrule
\end{tabular}
}
\end{table}

\subsection{Attack Effectiveness under Defense Mechanisms}
To investigate the robustness of TDOA under defense settings, we examine two mainstream defense methods: Self-reminder~\cite{xie2023defending} and Paraphrase~\cite{jain2023baseline}. 
The former augments the input by adding prompts before and after the text to caution LLMs about potential adversarial examples, 
while the latter leverages LLMs (GPT-4o) to rewrite the text while preserving its original meaning~\cite{jain2023baseline}. 
As shown in Table~\ref{def}, even after applying defenses, TDOA-1 achieves up to 32.51\% in LI-ASR and 46.64\% in ASR, 
demonstrating that TDOA remains effective against defended models.

\begin{table}[t]
\centering
\caption{LI-ASR (\%) and ASR (\%) under defense settings. 
``Original'' denotes the attack performance of TDOA-1 without defense, 
``Self-reminder'' denotes the performance with the Self-reminder defense, 
and ``Paraphrase'' denotes the performance with the Paraphrase defense.}
\label{def}
\resizebox{0.48\textwidth}{!}{%
\begin{tabular}{@{}c|c|ccc|cc@{}}
\toprule
\multirow{2}{*}{Datasets} & \multirow{2}{*}{With and without defense} & \multicolumn{3}{c|}{LI-ASR}    & \multicolumn{2}{c}{ASR} \\ \cmidrule(l){3-7} 
                          &                                           & Bert    & Distilbert & Roberta & GPT-4o     & GPT-4.1    \\ \midrule \midrule
\multirow{3}{*}{Emotion}  & Original                                  & 42.3\% & 52.9\%    & 41.6\% & 40.8\%    & 50.3\%    \\
                          & Self-reminder                             & - & -    & - & 34.0\%    & 45.6\%    \\
                          & Paraphrase                                & 28.10\% & 27.9\%    & 23.1\% & 21.4\%    & 36.0\%    \\ \midrule
\multirow{3}{*}{SST5}     & Original                                  & 32.11\% & 46.1\%    & 29.3\% & 32.0\%    & 50.8\%    \\
                          & Self-reminder                             & - & -    & - & 28.5\%    & 46.6\%    \\
                          & Paraphrase                                & 20.4\% & 32.5\%    & 21.8\% & 26.5\%    & 44.3\%    \\ \bottomrule
\end{tabular}
}
\end{table}

\section{Conclusion}

In this study, we introduce the concept of dynamic output scenarios, which encompass both multi-label classification tasks—where the number of output labels may vary across inputs—and generative models that produce inherently non-static outputs. With the rapid advancement of LLMs, such scenarios are becoming increasingly prevalent and demand systematic and comprehensive investigation.

To address the challenges posed by these scenarios, we propose a novel evaluation metric, termed \textbf{Label Intersection Attack Success Rate (LI-ASR)}. This metric is specifically designed to quantify the effectiveness of adversarial examples in multi-label output settings, thereby capturing the unique complexities that arise when outputs cannot be reduced to a single static label.

Beyond evaluation, we further contribute the first algorithm explicitly tailored for dynamic output contexts, referred to as the \textbf{TDOA attack method}. TDOA consists of two core components: surrogate model training and the farthest-label targeted attack strategy. The surrogate model training transforms dynamic output tasks into equivalent static formulations, enabling existing textual attack algorithms to be seamlessly applied in a plug-and-play manner. To further enhance transferability, particularly when discrepancies exist between surrogate and victim models, we incorporate the farthest-label targeted attack strategy. This train-free enhancement substantially amplifies the attack’s effectiveness without requiring modifications to model architectures or loss functions.

Overall, our contributions advance the study of adversarial robustness in dynamic output scenarios by (i) formalizing a new problem setting, (ii) introducing a principled evaluation metric, (iii) proposing the first dedicated attack algorithm, and (iv) demonstrating how these innovations improve both the efficiency and efficacy of adversarial attacks, while also informing the design of more robust defense strategies.
\bibliographystyle{IEEEtran}
\bibliography{IEEEabrv,reference}

\begin{thebibliography}{10}
\providecommand{\url}[1]{#1}
\csname url@samestyle\endcsname
\providecommand{\newblock}{\relax}
\providecommand{\bibinfo}[2]{#2}
\providecommand{\BIBentrySTDinterwordspacing}{\spaceskip=0pt\relax}
\providecommand{\BIBentryALTinterwordstretchfactor}{4}
\providecommand{\BIBentryALTinterwordspacing}{\spaceskip=\fontdimen2\font plus
\BIBentryALTinterwordstretchfactor\fontdimen3\font minus \fontdimen4\font\relax}
\providecommand{\BIBforeignlanguage}[2]{{%
\expandafter\ifx\csname l@#1\endcsname\relax
\typeout{** WARNING: IEEEtran.bst: No hyphenation pattern has been}%
\typeout{** loaded for the language `#1'. Using the pattern for}%
\typeout{** the default language instead.}%
\else
\language=\csname l@#1\endcsname
\fi
#2}}
\providecommand{\BIBdecl}{\relax}
\BIBdecl

\bibitem{wei2018transferable}
X.~Wei, S.~Liang, N.~Chen, and X.~Cao, ``Transferable adversarial attacks for image and video object detection,'' \emph{arXiv preprint arXiv:1811.12641}, 2018.

\bibitem{liang2020efficient}
S.~Liang, X.~Wei, S.~Yao, and X.~Cao, ``Efficient adversarial attacks for visual object tracking,'' in \emph{Computer Vision--ECCV 2020: 16th European Conference, Glasgow, UK, August 23--28, 2020, Proceedings, Part XXVI 16}, 2020.

\bibitem{liang2022large}
S.~Liang, L.~Li, Y.~Fan, X.~Jia, J.~Li, B.~Wu, and X.~Cao, ``A large-scale multiple-objective method for black-box attack against object detection,'' in \emph{European Conference on Computer Vision}, 2022.

\bibitem{liang2022parallel}
S.~Liang, B.~Wu, Y.~Fan, X.~Wei, and X.~Cao, ``Parallel rectangle flip attack: A query-based black-box attack against object detection,'' \emph{arXiv preprint arXiv:2201.08970}, 2022.

\bibitem{muxue2023adversarial}
L.~Muxue, C.~Wang, S.~Liang, A.~Liu, Z.~Liu, L.~Yang, and X.~Cao, ``Adversarial instance attacks for interactions between human and object.''

\bibitem{wang2023diversifying}
Z.~Wang, Z.~Zhang, S.~Liang, and X.~Wang, ``Diversifying the high-level features for better adversarial transferability,'' \emph{arXiv preprint arXiv:2304.10136}, 2023.

\bibitem{liu2023x}
A.~Liu, J.~Guo, J.~Wang, S.~Liang, R.~Tao, W.~Zhou, C.~Liu, X.~Liu, and D.~Tao, ``$\{$X-Adv$\}$: Physical adversarial object attacks against x-ray prohibited item detection,'' in \emph{32nd USENIX Security Symposium (USENIX Security 23)}, 2023.

\bibitem{waghela2024modified}
H.~Waghela, S.~Rakshit, and J.~Sen, ``A modified word saliency-based adversarial attack on text classification models,'' \emph{arXiv preprint arXiv:2403.11297}, 2024.

\bibitem{han2024bfs2adv}
X.~Han, Q.~Li, H.~Cao, L.~Han, B.~Wang, X.~Bao, Y.~Han, and W.~Wang, ``Bfs2adv: Black-box adversarial attack towards hard-to-attack short texts,'' \emph{CS}, p. 103817, 2024.

\bibitem{zhu2024limeattack}
H.~Zhu, Q.~Zhao, W.~Shang, Y.~Wu, and K.~Liu, ``Limeattack: Local explainable method for textual hard-label adversarial attack,'' in \emph{Proceedings of the AAAI Conference on Artificial Intelligence}, vol.~38, no.~17, 2024, pp. 19\,759--19\,767.

\bibitem{kang2024hybrid}
Y.~Kang, J.~Zhao, X.~Yang, B.~Fan, and W.~Xie, ``A hybrid style transfer with whale optimization algorithm model for textual adversarial attack,'' \emph{NCA}, vol.~36, pp. 4263--4280, 2024.

\bibitem{kassim2024multi}
M.~A. Kassim, H.~Viktor, and W.~Michalowski, ``Multi-label lifelong machine learning: A scoping review of algorithms, techniques, and applications,'' \emph{IEEE Access}, 2024.

\bibitem{fan2024learning}
Y.~Fan, J.~Liu, J.~Tang, P.~Liu, Y.~Lin, and Y.~Du, ``Learning correlation information for multi-label feature selection,'' \emph{Pattern Recognition}, vol. 145, p. 109899, 2024.

\bibitem{lu2025adversarial}
L.~Lu, S.~Pang, S.~Liang, H.~Zhu, X.~Zeng, A.~Liu, Y.~Liu, and Y.~Zhou, ``Adversarial training for multimodal large language models against jailbreak attacks,'' \emph{arXiv preprint arXiv:2503.04833}, 2025.

\bibitem{liu2023exploring}
A.~Liu, S.~Tang, S.~Liang, R.~Gong, B.~Wu, X.~Liu, and D.~Tao, ``Exploring the relationship between architectural design and adversarially robust generalization,'' in \emph{Proceedings of the IEEE/CVF Conference on Computer Vision and Pattern Recognition}, 2023.

\bibitem{wang2022semattack}
B.~Wang, C.~Xu, X.~Liu, Y.~Cheng, and B.~Li, ``Semattack: Natural textual attacks via different semantic spaces,'' in \emph{NAACL}, 2022, pp. 176--205.

\bibitem{hu2024fasttextdodger}
X.~Hu, G.~Liu, B.~Zheng, L.~Zhao, Q.~Wang, Y.~Zhang, and M.~Du, ``Fasttextdodger: Decision-based adversarial attack against black-box nlp models with extremely high efficiency,'' \emph{TIFS}, 2024.

\bibitem{liu2024hqa}
H.~Liu, Z.~Xu, X.~Zhang, F.~Zhang, F.~Ma, H.~Chen, H.~Yu, and X.~Zhang, ``Hqa-attack: Toward high quality black-box hard-label adversarial attack on text,'' \emph{NeurIPS}, vol.~36, 2024.

\bibitem{liu2023sspattack}
H.~Liu, Z.~Xu, X.~Zhang, X.~Xu, F.~Zhang, F.~Ma, H.~Chen, H.~Yu, and X.~Zhang, ``Sspattack: a simple and sweet paradigm for black-box hard-label textual adversarial attack,'' in \emph{AAAI}, vol.~37, no.~11, 2023, pp. 13\,228--13\,235.

\bibitem{lin2021using}
J.~Lin, J.~Zou, and N.~Ding, ``Using adversarial attacks to reveal the statistical bias in machine reading comprehension models,'' in \emph{ACL}, 2021, pp. 333--342.

\bibitem{xu2021grey}
Y.~Xu, X.~Zhong, A.~J. Yepes, and J.~H. Lau, ``Grey-box adversarial attack and defence for sentiment classification,'' in \emph{ACL}, 2021, pp. 4078--4087.

\bibitem{wang2020t3}
B.~Wang, H.~Pei, B.~Pan, Q.~Chen, S.~Wang, and B.~Li, ``T3: Tree-autoencoder constrained adversarial text generation for targeted attack,'' in \emph{EMNLP}, 2020, pp. 6134--6150.

\bibitem{le2020malcom}
T.~Le, S.~Wang, and D.~Lee, ``Malcom: Generating malicious comments to attack neural fake news detection models,'' in \emph{ICDM}, 2020, pp. 282--291.

\bibitem{li2019textbugger}
J.~Li, S.~Ji, T.~Du, B.~Li, and T.~Wang, ``Textbugger: Generating adversarial text against real-world applications,'' in \emph{NDSS}, 2019.

\bibitem{lee2022query}
D.~Lee, S.~Moon, J.~Lee, and H.~O. Song, ``Query-efficient and scalable black-box adversarial attacks on discrete sequential data via bayesian optimization,'' in \emph{ICML}, 2022, pp. 12\,478--12\,497.

\bibitem{li2020Bert}
L.~Li, R.~Ma, Q.~Guo, X.~Xue, and X.~Qiu, ``Bert-attack: Adversarial attack against bert using bert,'' \emph{arXiv e-prints}, pp. arXiv--2004, 2020.

\bibitem{zang2020word}
Y.~Zang, F.~Qi, C.~Yang, Z.~Liu, M.~Zhang, Q.~Liu, and M.~Sun, ``Word-level textual adversarial attacking as combinatorial optimization,'' in \emph{ACL}, 2020, pp. 6066--6080.

\bibitem{papernot2017practical}
N.~Papernot, P.~McDaniel, I.~Goodfellow, S.~Jha, Z.~B. Celik, and A.~Swami, ``Practical black-box attacks against machine learning,'' in \emph{CCS}, 2017, pp. 506--519.

\bibitem{dong2018boosting}
Y.~Dong, F.~Liao, T.~Pang, H.~Su, J.~Zhu, X.~Hu, and J.~Li, ``Boosting adversarial attacks with momentum,'' in \emph{CVPR}, 2018, pp. 9185--9193.

\bibitem{li2020practical}
Q.~Li, Y.~Guo, and H.~Chen, ``Practical no-box adversarial attacks against dnns,'' \emph{NeurIPS}, vol.~33, pp. 12\,849--12\,860, 2020.

\bibitem{sun2022towards}
C.~Sun, Y.~Zhang, W.~Chaoqun, Q.~Wang, Y.~Li, T.~Liu, B.~Han, and X.~Tian, ``Towards lightweight black-box attack against deep neural networks,'' \emph{NeurIPS}, vol.~35, pp. 19\,319--19\,331, 2022.

\bibitem{wang2021feature}
Z.~Wang, H.~Guo, Z.~Zhang, W.~Liu, Z.~Qin, and K.~Ren, ``Feature importance-aware transferable adversarial attacks,'' in \emph{ICCV}.\hskip 1em plus 0.5em minus 0.4em\relax IEEE, 2021, pp. 7619--7628.

\bibitem{richards2021adversarial}
L.~E. Richards, A.~Nguyen, R.~Capps, S.~Forsyth, C.~Matuszek, and E.~Raff, ``Adversarial transfer attacks with unknown data and class overlap,'' in \emph{ACM}, 2021, pp. 13--24.

\bibitem{xiaosen2023rethinking}
W.~Xiaosen, K.~Tong, and K.~He, ``Rethinking the backward propagation for adversarial transferability,'' \emph{NeurIPS}, vol.~36, pp. 1905--1922, 2023.

\bibitem{yuan2021meta}
Z.~Yuan, J.~Zhang, Y.~Jia, C.~Tan, T.~Xue, and S.~Shan, ``Meta gradient adversarial attack,'' in \emph{ICCV}.\hskip 1em plus 0.5em minus 0.4em\relax IEEE, 2021, pp. 7728--7737.

\bibitem{lee2023gelu}
M.~Lee, ``Gelu activation function in deep learning: a comprehensive mathematical analysis and performance,'' \emph{arXiv preprint arXiv:2305.12073}, 2023.

\bibitem{zhou2024towards}
P.~Zhou, X.~Xie, Z.~Lin, and S.~Yan, ``Towards understanding convergence and generalization of adamw,'' \emph{TPAMI}, 2024.

\bibitem{wang2025no}
W.~Wang, S.~Liang, Y.~Zhang, X.~Jia, H.~Lin, and X.~Cao, ``No query, no access,'' \emph{arXiv preprint arXiv:2505.07258}, 2025.

\bibitem{hsu2017hybrid}
S.~T. Hsu, C.~Moon, P.~Jones, and N.~Samatova, ``A hybrid cnn-rnn alignment model for phrase-aware sentence classification,'' in \emph{EACL}, 2017, pp. 443--449.

\bibitem{saravia-etal-2018-carer}
E.~Saravia, H.-C.~T. Liu, Y.-H. Huang, J.~Wu, and Y.-S. Chen, ``{CARER}: Contextualized affect representations for emotion recognition,'' in \emph{EMNLP}, 2018, pp. 3687--3697.

\bibitem{formento2025confidence}
B.~Formento, C.~S. Foo, and S.~K. Ng, ``Confidence elicitation: A new attack vector for large language models,'' in \emph{Proceedings of the Thirteenth International Conference on Learning Representations (ICLR)}, 2025.

\bibitem{zhang2025emotiattack}
Y.~Zhang, ``Emoti-attack: Zero-perturbation adversarial attacks on nlp systems via emoji sequences,'' \emph{arXiv preprint arXiv:2502.17392}, 2025.

\bibitem{ye2022leapattack}
M.~Ye, J.~Chen, C.~Miao, T.~Wang, and F.~Ma, ``Leapattack: Hard-label adversarial attack on text via gradient-based optimization,'' in \emph{SIGKDD}, 2022, pp. 2307--2315.

\bibitem{garg2020Bae}
S.~Garg and G.~Ramakrishnan, ``Bae: Bert-based adversarial examples for text classification,'' in \emph{EMNLP}, 2020, pp. 6174--6181.

\bibitem{lv2023ct}
M.~Lv, C.~Dai, K.~Li, W.~Zhou, and S.~Hu, ``Ct-gat: Cross-task generative adversarial attack based on transferability,'' in \emph{EMNLP}, 2024.

\bibitem{gao2018black}
J.~Gao, J.~Lanchantin, M.~L. Soffa, and Y.~Qi, ``Black-box generation of adversarial text sequences to evade deep learning classifiers,'' in \emph{SPW}, 2018, pp. 50--56.

\bibitem{papernot2016crafting}
N.~Papernot, P.~McDaniel, A.~Swami, and R.~Harang, ``Crafting adversarial input sequences for recurrent neural networks,'' in \emph{MILCOM}, 2016, pp. 49--54.

\bibitem{ren2019generating}
S.~Ren, Y.~Deng, K.~He, and W.~Che, ``Generating natural language adversarial examples through probability weighted word saliency,'' in \emph{ACL}, 2019, pp. 1085--1097.

\bibitem{ahmed2020k}
M.~Ahmed, R.~Seraj, and S.~M.~S. Islam, ``The k-means algorithm: A comprehensive survey and performance evaluation,'' \emph{Electronics}, vol.~9, no.~8, p. 1295, 2020.

\bibitem{ni2022sentence}
J.~Ni, G.~H. Abrego, N.~Constant, J.~Ma, K.~Hall, D.~Cer, and Y.~Yang, ``Sentence-t5: Scalable sentence encoders from pre-trained text-to-text models,'' in \emph{ACL}, 2022, pp. 1864--1874.

\bibitem{lei2024prompt}
Y.~Lei, J.~Li, Z.~Li, Y.~Cao, and H.~Shan, ``Prompt learning in computer vision: a survey,'' \emph{Frontiers of Information Technology \& Electronic Engineering}, vol.~25, no.~1, pp. 42--63, 2024.

\bibitem{ho2025analisis}
J.~Ho, D.~F. Ramadhan, and A.~R. Aluska, ``Analisis peran gen ai dalam penetration testing: Studi kasus mesin vulnhub menggunakan gpt-4.1 dan kali linux,'' \emph{Bridge: Jurnal Publikasi Sistem Informasi dan Telekomunikasi}, vol.~3, no.~3, pp. 38--71, 2025.

\bibitem{hurst2024gpt}
A.~Hurst, A.~Lerer, A.~P. Goucher, A.~Perelman, A.~Ramesh, A.~Clark, A.~Ostrow, A.~Welihinda, A.~Hayes, A.~Radford \emph{et~al.}, ``Gpt-4o system card,'' \emph{arXiv preprint arXiv:2410.21276}, 2024.

\bibitem{radford2021learning}
A.~Radford, J.~W. Kim, C.~Hallacy, A.~Ramesh, G.~Goh, S.~Agarwal, G.~Sastry, A.~Askell, P.~Mishkin, J.~Clark \emph{et~al.}, ``Learning transferable visual models from natural language supervision,'' in \emph{ICML}, 2021, pp. 8748--8763.

\bibitem{yang2019xlnet}
Z.~Yang, Z.~Dai, Y.~Yang, J.~Carbonell, R.~R. Salakhutdinov, and Q.~V. Le, ``Xlnet: Generalized autoregressive pretraining for language understanding,'' \emph{NIPS}, vol.~32, 2019.

\bibitem{okada2019efficient}
S.~Okada, M.~Ohzeki, and S.~Taguchi, ``Efficient partition of integer optimization problems with one-hot encoding,'' \emph{Scientific reports}, vol.~9, no.~1, p. 13036, 2019.

\bibitem{zhang1996birch}
T.~Zhang, R.~Ramakrishnan, and M.~Livny, ``Birch: an efficient data clustering method for very large databases,'' \emph{ACM sigmod record}, vol.~25, no.~2, pp. 103--114, 1996.

\bibitem{von2007tutorial}
U.~Von~Luxburg, ``A tutorial on spectral clustering,'' \emph{Statistics and computing}, vol.~17, pp. 395--416, 2007.

\bibitem{baidu2019translate}
Baidu, ``Baidu translate,'' 2019, accessed: 2023-10-31.

\bibitem{alibaba2020translate}
A.~Group, ``Alibaba cloud machine translation,'' 2020, accessed: 2023-10-31.

\bibitem{huang2010transcription}
C.~S. Huang, \emph{A transcription for piano and oboe of the Sonata for Piano and Violin in A major, Opus 100, by Johannes Brahms, with commentary on analysis and methodology}.\hskip 1em plus 0.5em minus 0.4em\relax University of Hartford, 2010.

\bibitem{chertkov2023translate}
A.~Chertkov, O.~Tsymboi, M.~Pautov, and I.~Oseledets, ``Translate your gibberish: black-box adversarial attack on machine translation systems,'' \emph{arXiv preprint arXiv:2303.10974}, 2023.

\bibitem{sadrizadehtransfool}
S.~Sadrizadeh, L.~Dolamic, and P.~Frossard, ``Transfool: An adversarial attack against neural machine translation models,'' \emph{TMLR}, 2023.

\bibitem{tan-etal-2020-morphin}
S.~Tan, S.~Joty, M.-Y. Kan, and R.~Socher, ``It{'}s morphin{'} time! {C}ombating linguistic discrimination with inflectional perturbations,'' in \emph{ACL}, 2020, pp. 2920--2935.

\bibitem{papineni2002bleu}
K.~Papineni, S.~Roukos, T.~Ward, and W.-J. Zhu, ``Bleu: a method for automatic evaluation of machine translation,'' in \emph{ACL}, 2002, pp. 311--318.

\bibitem{popovic2015chrf}
M.~Popovi{\'c}, ``chrf: character n-gram f-score for automatic mt evaluation,'' in \emph{SMT}, 2015, pp. 392--395.

\bibitem{xie2023defending}
Y.~Xie, J.~Yi, J.~Shao, J.~Curl, L.~Lyu, Q.~Chen, X.~Xie, and F.~Wu, ``Defending chatgpt against jailbreak attack via self-reminders,'' \emph{Nature Machine Intelligence}, vol.~5, no.~12, pp. 1486--1496, 2023.

\bibitem{jain2023baseline}
N.~Jain, A.~Schwarzschild, Y.~Wen, G.~Somepalli, J.~Kirchenbauer, P.-Y. Chiang, M.~Goldblum, A.~Saha, J.~Geiping, and T.~Goldstein, ``Baseline defenses for adversarial attacks against aligned language models,'' \emph{arXiv preprint arXiv:2309.00614}, 2023.

\end{thebibliography}

\end{document}